\documentclass[lettersize,journal]{IEEEtran}
\usepackage{amsmath,amsfonts}
\usepackage{algorithmic}
\usepackage{algorithm}
\usepackage[caption=false,font=normalsize,labelfont=sf,textfont=sf]{subfig}
\usepackage{booktabs}
\usepackage{multirow}
\usepackage{array}
\usepackage{caption}
\usepackage{nicematrix}
\usepackage{textcomp}
\usepackage{stfloats}
\usepackage{url}
\usepackage{ragged2e}
\usepackage{verbatim}
\usepackage{graphicx}
\usepackage{cite}
\usepackage{tikz}
\usepackage{pgfplots}
\pgfplotsset{compat=1.18}  
\usepackage{graphicx}
\usepackage{tabularx}   
\usepackage{booktabs}
\usepackage{adjustbox}
\usepackage[table]{xcolor}
\usepackage{colortbl}
\usepackage[colorlinks=true, linkcolor=blue, citecolor=blue, urlcolor=magenta]{hyperref}
\usepackage{orcidlink}
\usepackage{balance}

\begin{document}
\newcolumntype{P}[1]{>{\arraybackslash}p{#1}}
\captionsetup{justification=justified, singlelinecheck=false}
\captionsetup[table]{labelsep=newline, justification=centering}
\renewcommand{\figurename}{Fig.}
\captionsetup[figure]{labelsep=period}

\title{Towards Next-Generation SLAM: A Survey on 3DGS-SLAM Focusing on Performance, Robustness, and Future Directions}

\author{
Li Wang\href{https://orcid.org/0000-0002-9325-2391}{\orcidlink{0000-0002-9325-2391}},
Ruixuan Gong\href{https://orcid.org/0009-0009-1303-0982}{\orcidlink{0009-0009-1303-0982}},
Yumo Han\href{https://orcid.org/0009-0006-9881-8979}{\orcidlink{0009-0006-9881-8979}},
Lei Yang\href{https://orcid.org/0000-0003-1800-6892}{\orcidlink{0000-0003-1800-6892}}, \textit{Member, IEEE},
Lu Yang\href{https://orcid.org/0009-0007-3506-7079}{\orcidlink{0009-0007-3506-7079}},
Ying Li\href{https://orcid.org/0000-0003-0608-9619}{\orcidlink{0000-0003-0608-9619}},
Bin Xu\href{https://orcid.org/0000-0001-8934-3074}{\orcidlink{0000-0001-8934-3074}},
Huaping Liu\href{https://orcid.org/0000-0002-4042-6044}{\orcidlink{0000-0002-4042-6044}}, \textit{Fellow, IEEE},
and Rong Fu \href{https://orcid.org/0009-0001-4549-8119}{\orcidlink{0009-0001-4549-8119}}

\thanks{This work was supported by the National Natural Science Foundation of China under Grant No. 52502496, U22B2052 and the Natural Science Foundation of Chongqing, China under Grant No. CSTB2025NSCQ-GPX0413, and the National High Technology Research and Development Program of China under Grant No. 2020YFC1512501. \textit{(Corresponding author: Rong Fu.)}}
\thanks{Li Wang is with School of Mechanical Engineering, Beijing Institute of Technology, Beijing 100081, China and Chongqing Innovation Center, Beijing Institute of Technology, Chongqing 401120, China (e-mail: wangli\_bit@bit.edu.cn).}
\thanks{Ruixuan Gong, Lu Yang, Ying Li and Bin Xu are with School of Mechanical Engineering, Beijing Institute of Technology, Beijing 100081, China (e-mail: 3220240420@bit.edu.cn, yanglu@bit.edu.cn, ying.li@bit.edu.cn, bitxubin@bit.edu.cn).}
\thanks{Yumo Han is with the School of Artificial Intelligence, University of Science and Technology Beijing, Beijing 100083, China (e-mail: hym2004227@163.com).}
\thanks{Lei Yang is with the School of Mechanical and Aerospace Engineering, Nanyang Technological University, Singapore 639798, Singapore (e-mail: lei.yang@ntu.edu.sg).}
\thanks{Huaping Liu is with the State Key Laboratory of Intelligent Technology and Systems, and the Department of Computer Science and Technology, Tsinghua University, Beijing 100084, China (e-mail: hpliu@tsinghua.edu.cn).}
\thanks{Rong Fu is with the Shanghai AI laboratory, Shanghai 200003, China (e-mail: furong@pjlab.org.cn).}
}



\maketitle

\begin{abstract}
Traditional Simultaneous Localization and Mapping (SLAM) systems often face limitations including coarse rendering quality, insufficient recovery of scene details, and poor robustness in dynamic environments. 3D Gaussian Splatting (3DGS), with its efficient explicit representation and high-quality rendering capabilities, offers a new reconstruction paradigm for SLAM. This survey comprehensively reviews key technical approaches for integrating 3DGS with SLAM. We analyze performance optimization of representative methods across four critical dimensions: rendering quality, tracking accuracy, reconstruction speed, and memory consumption, delving into their design principles and breakthroughs. Furthermore, we examine methods for enhancing the robustness of 3DGS-SLAM in complex environments such as motion blur and dynamic environments. Finally, we discuss future challenges and development trends in this area. This survey aims to provide a technical reference for researchers and foster the development of next-generation SLAM systems characterized by high fidelity, efficiency, and robustness.
\end{abstract}


\section{Introduction}
\IEEEPARstart{S}{imultaneous} Localization and Mapping (SLAM) serves as a fundamental technology for applications including autonomous navigation, augmented reality, and autonomous driving, aiming to estimate the pose within an unknown environment while simultaneously constructing a consistent map. SLAM fuses sensor data to extract environmental features and iteratively optimize both pose and map based on motion and observation models.

Traditional SLAM methods rely mainly on geometric features, evolving from filter approaches (e.g., EKF-SLAM\cite{EKF-SLAM}) to sparse feature point methods (e.g., MonoSLAM\cite{MonoSLAM} and PTAM\cite{PTAM}), then to dense/semi-dense methods (e.g., DTAM\cite{DTAM} and LSD-SLAM\cite{LSD_SLAM}), and finally to tightly coupled architectures (e.g., the ORB-SLAM series\cite{ORB-SLAM3} and DSO\cite{DSO}). MonoSLAM and PTAM pioneered real-time parallel tracking, while direct methods like LSD-SLAM and DSO improved robustness in low-texture environments. Furthermore, the ORB-SLAM series established the classic graph-based paradigm by integrating loop closure and rigorous keyframe management. With the development of deep learning, learning-based methods (e.g., DROID-SLAM\cite{DROID_SLAM}, DeepFactors\cite{DeepFactors} and SP-SLAM\cite{SP-SLAM}) have significantly improved accuracy. Methods that integrate semantic information into mapping (e.g., MaskFusion\cite{MaskFusion}, Co-Fusion\cite{Co-Fusion} and RDS-SLAM\cite{RDS-SLAM}) have further enhanced the ability to understand and model the environment\cite{PAS-SLAM}.

Despite their advantages, these methods still face notable limitations. In terms of rendering quality, many methods rely on coarse geometric representations (e.g., sparse point clouds and meshes) that recover only rough scene geometry and cannot generate photorealistic views. In tracking accuracy, feature matching can fail under weak textures, dynamic motion, or lighting changes, causing pose drift. Additionally, memory consumption remains a challenge, as dense methods often store meshes or voxels, leading to high memory usage.

Recently, Neural Radiance Fields (NeRF) \cite{NeRF} and its variants\cite{SPIn-NeRF,Mip-NeRF,instantngp,Depth-guided,Dynamic-NeRF,I-DACS} have contributed to significant advancements in SLAM. Systems like iMAP\cite{iMAP}, Nice-SLAM\cite{NICE-SLAM}, ESLAM\cite{ESLAM}, and Point-SLAM\cite{Point-SLAM} combine NeRF’s high-fidelity reconstruction with SLAM’s pose optimization to enable learnable, differentiable, end-to-end mapping. Seminal works like iMAP and NICE-SLAM demonstrated the potential of implicit fields for continuous reconstruction. Subsequent methods introduced more efficient representations, such as the axis-aligned feature planes in ESLAM or the neural point cloud representations in Point-SLAM. However, NeRF-based methods rely on dense view sampling and struggle with sparse views. Neural network training is computationally expensive, making real-time requirements difficult to meet. These methods also tend to use large voxel hashes or multi-layer perceptrons (MLPs) for scene representation, resulting in high complexity.

\begin{figure*}[!t]
  \centering
  \includegraphics[width=\textwidth]{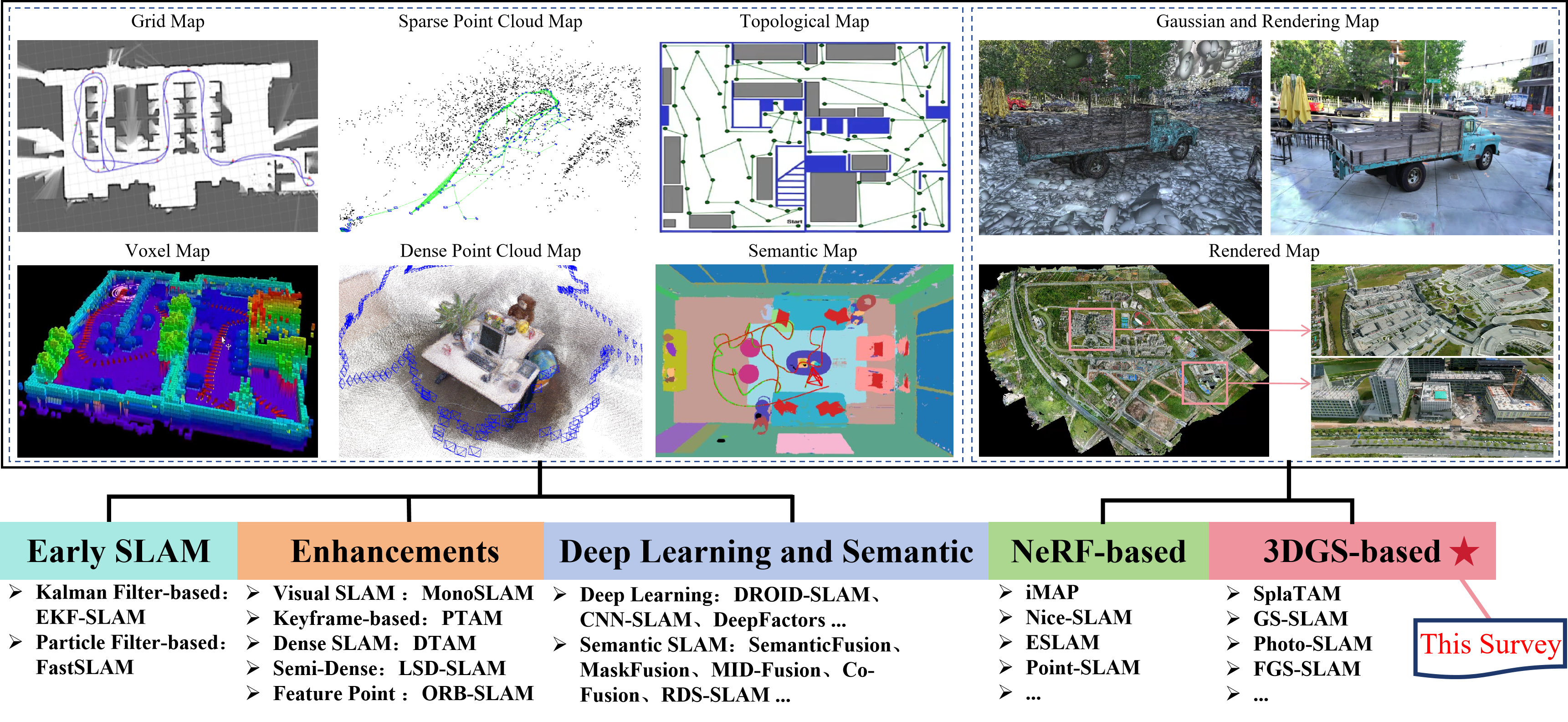} 
  \caption{Typical SLAM map representations and evolution of SLAM. The upper subfigures display diverse scene representations enabled by various SLAM approaches, highlighting the transition from simple geometric reconstructions to rich, visually realistic scene models. The lower panel presents the evolutionary stages of SLAM: starting from early probabilistic filters, through keyframe and feature-based enhancements, to the integration of deep learning and semantic reasoning, the recent adoption of NeRF, and finally the latest 3DGS approaches which are the focus of this survey.}
  \label{1}
\end{figure*}

Existing SLAM paradigms face a fundamental trade-off between high-fidelity reconstruction and real-time efficiency. While explicit geometric representations (e.g., point clouds, voxels, meshes) facilitate real-time operation, they often fail to capture high-frequency texture details. Conversely, implicit neural representations excel at detail synthesis but incur prohibitive computational and memory costs, limiting their real-time deployability. The emerging 3D Gaussian Splatting (3DGS) technique \cite{3DGS} bridges this gap by offering an explicit representation that combines the rendering quality of NeRF with exceptional rendering speed. Although 3DGS and its variants \cite{FlashGS,VastGaussian,Taming3DGS,gsplat,Scaffold-gs,StreetSurfGS,GET3DGS,GEDR,AdR-Gaussian,grendel-gs,DashGaussian} significantly outperform NeRF in efficiency, the original framework was designed for offline optimization with known poses, restricting its direct application in online scenarios. Consequently, researchers have begun to integrate 3DGS into SLAM pipelines, combining robust real-time pose estimation with high-quality scene reconstruction. This synergy establishes a new generation of visual SLAM capable of achieving simultaneous high-fidelity and real-time mapping. Fig.~\ref{1} shows the development of SLAM.

Since the introduction of 3DGS, several surveys\cite{z5,z6,z7,z8,z9,z10} have reviewed it, but focus on 3DGS as a general representation of the scene without exploring the specific optimization challenges when integrating it with SLAM. Conversely, existing SLAM surveys\cite{z12,z13,z14,z15} do not address the potential of 3DGS in SLAM. Some works\cite{z17,z18,z19} have attempted to summarize the progress of 3DGS-SLAM, but these typically categorize it based on traditional SLAM (e.g., by sensor modality), neglecting the core requirements across different applications. For instance, immersive AR/VR requires high consistency between virtual overlays and the real world, demanding excellent rendering quality; autonomous robotics\cite{SAT-GCN} and UAV navigation\cite{BEVHeight} require stable pose estimation for safety, needing enhanced tracking accuracy; autonomous driving\cite{Dual} and interactive digital twins rely on low latency, demanding optimized speed; large-scale mapping must handle massive data, highlighting the importance of memory optimization.

\begin{figure*}[!t]
  \centering
  \includegraphics[width=\textwidth]{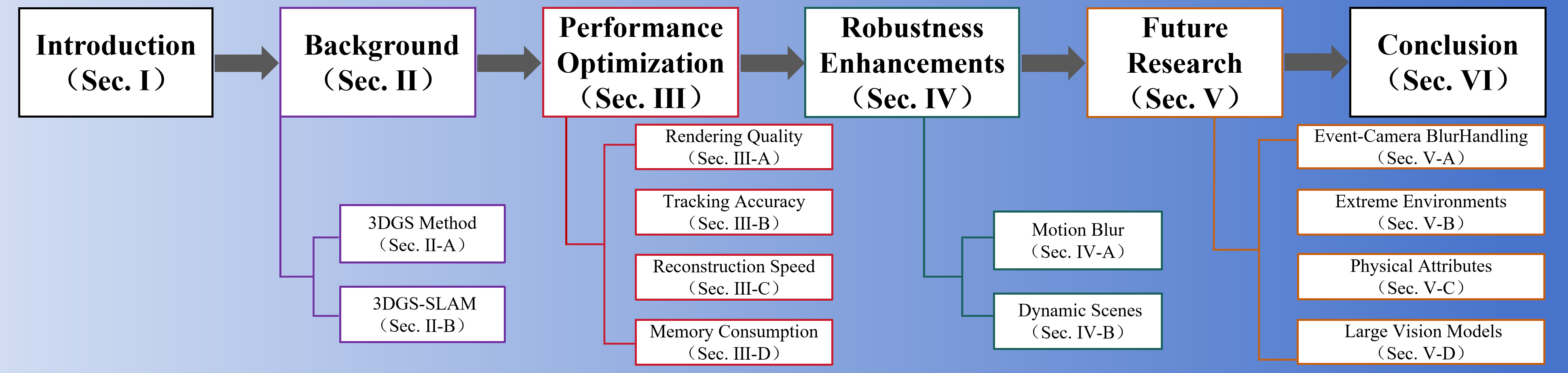} 
  \caption{Overall structure of the article.}
  \label{2}
\end{figure*}

Based on this perspective, this survey focuses on the optimization strategies for 3DGS-SLAM. We systematically examine core techniques and representative works in four key performance dimensions: rendering quality, tracking accuracy, reconstruction speed, and memory consumption. Additionally, we discuss methods for enhancing robustness in handling motion blur and dynamic scenes. Fig.~\ref{2} outlines the structure of this article. Our goal is to provide a comprehensive reference and facilitate the development of next-generation SLAM characterized by high fidelity, efficiency, and robustness.

\section{Background}
\subsection{3D Gaussian Splatting Method}
\begin{figure*}[!t]
  \centering
  \includegraphics[width=\textwidth]{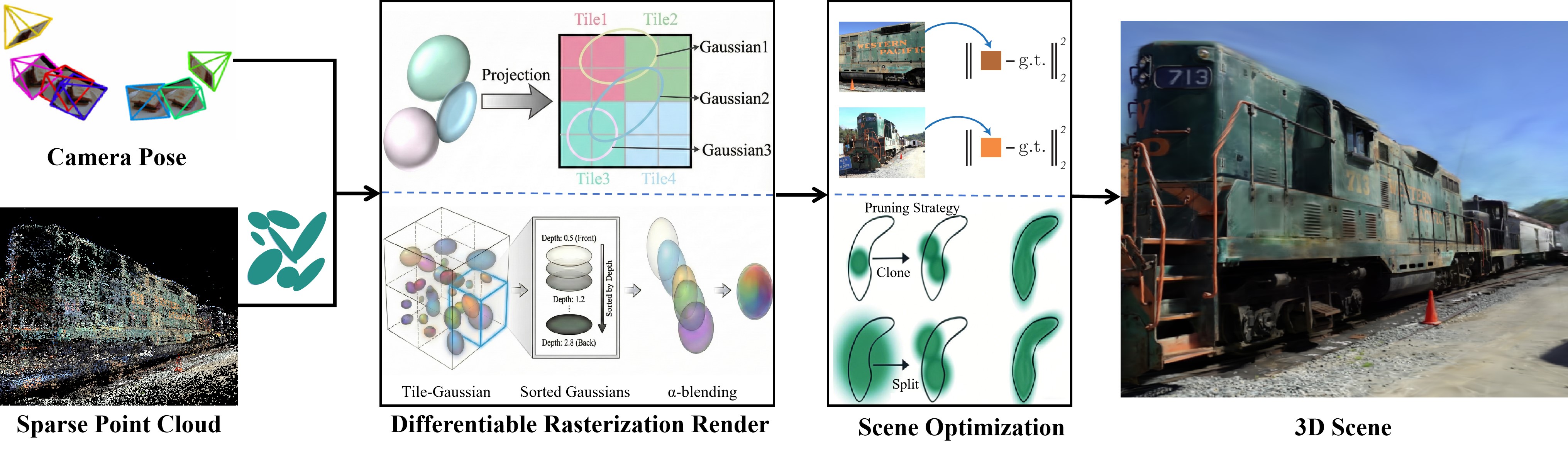} 
  \caption{General pipeline of 3D Gaussian Splatting. Initialized from sparse points, the method renders views via differentiable rasterization and iteratively refines the geometry through adaptive optimization.}
  \label{3}
\end{figure*}

The 3DGS framework encompasses four core algorithmic stages: \textbf{point cloud and Gaussian primitive initialization}, \textbf{differentiable projection}, \textbf{rasterized rendering}, and \textbf{scene optimization}. The following sections analyze each stage in sequence to systematically delineate the overall pipeline. Fig.~\ref{3} illustrates the general 3DGS pipeline.

\textit{1)Point Cloud and Gaussian Primitive Initialization:}
A 3DGS system takes as input multi-view images and corresponding camera poses, often using structure-from-motion (SfM) to generate a sparse point cloud $\{\mathbf{p}_i\}$ as initialization. From this point cloud, each 3D Gaussian splat $G_i$ is initialized with parameters: position $\mu_i$, opacity $\alpha_i$, covariance $\Sigma_i$, and color $\mathbf{c}_i$(color is typically represented by spherical harmonics). The spatial density of a Gaussian is defined as
\begin{equation}
G_i(\mathbf{x}) = \exp\left(-\tfrac{1}{2} (\mathbf{x} - \mu_i)^\top \Sigma_i^{-1} (\mathbf{x} - \mu_i)\right).
\end{equation}
To ensure $\Sigma_i$ is positive semi-definite, it is reparameterized via a rotation $R_i$ and scale matrix $S_i$:
\begin{equation}
\Sigma_i = R_i S_i S_i^\top R_i^\top,
\end{equation}
where $S_i = \mathrm{diag}(s_{ix},s_{iy},s_{iz})$ and $R_i$ is generated from a learnable quaternion $q_i=(q_w,q_x,q_y,q_z)$.

\textit{2)Differentiable Projection:}
Given a camera pose, the 3DGS system first prunes Gaussians lying outside the view frustum. The remaining 3D Gaussians are then projected into the 2D image plane. Given a view transformation matrix $W$, the projected 2D center $\mu'$ and covariance $\Sigma'$ of each Gaussian are computed as:
\begin{equation}
    \mu' = W\mu,
\end{equation}
\begin{equation}
\Sigma' = J W \Sigma W^\top J^\top,
\end{equation}
where $J$ is the affine Jacobian of the projection.

\textit{3)Rasterized Rendering:}
For rendering, 3DGS uses a tile-based parallel rasterization\cite{Mesh-aligned} to avoid costly per-pixel iteration. The image is divided into non-overlapping $16 \times 16$ pixel tiles, and for each tile the system identifies which Gaussians project onto it.

Each tile is then processed in parallel: Gaussians are depth-sorted per tile to form an ordered list. Since tiles and pixels are independent, this approach is efficiently parallelized on CUDA. The color $C$ of a pixel is obtained by front-to-back alpha blending of the projected Gaussians:
\begin{equation}
C = \sum_{i \in N} c_i \alpha_i' T_i,
\end{equation}
where $N$ indexes Gaussians affecting the pixel,  $\alpha_i'$ is the effective opacity of Gaussian $i$ at the pixel, and $T_i$ is the cumulative transparency from preceding Gaussians. The effective opacity and transparency product are given by
\begin{equation}
\alpha_i' = \alpha_i \exp\left(-\frac{1}{2}(\mathbf{x}' - \boldsymbol{\mu}_i')^\top \Sigma_i'^{-1}(\mathbf{x}' - \boldsymbol{\mu}_i')\right),
\end{equation}
\begin{equation}
T_i=\prod_{j=1}^{i-1} (1 - \alpha_j').
\end{equation}
Gaussians with $\alpha_i'\textless1/255$ are discarded, and once $T_i$ falls below a threshold, further contributions are skipped, yielding the final pixel color.

\begin{figure*}[]
  \centering
  \includegraphics[width=\textwidth]{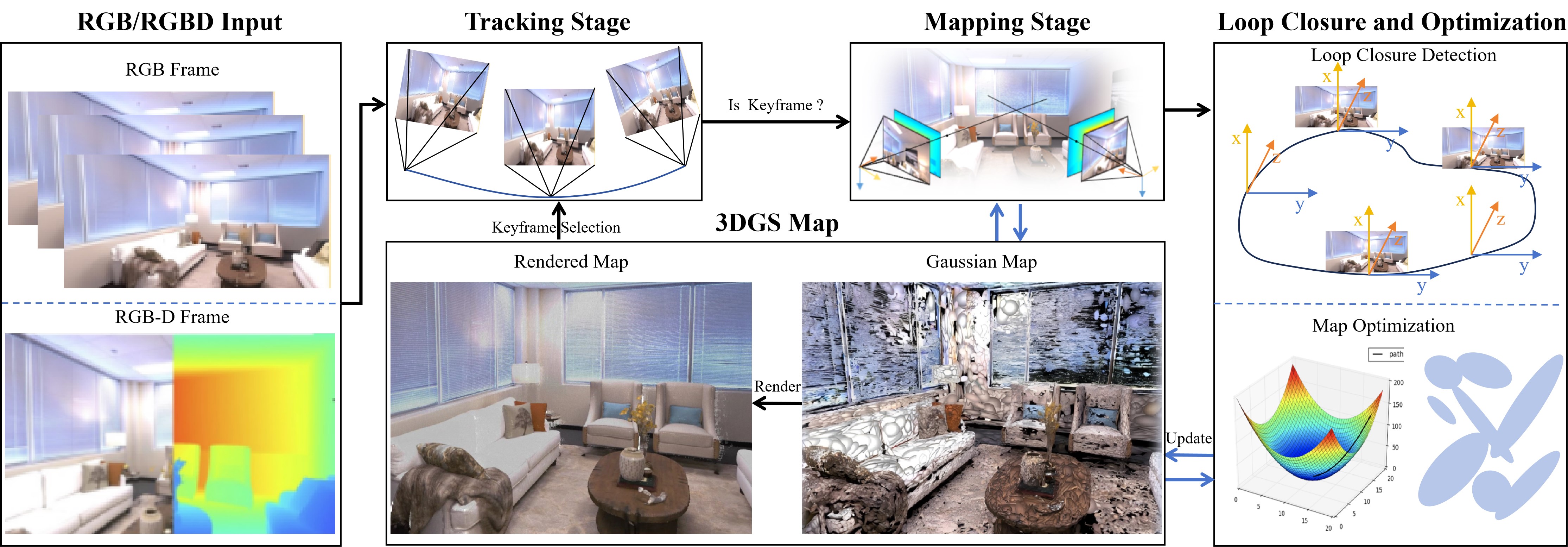} 
  \caption{General Pipeline of 3DGS-SLAM. Taking frames as input, the system performs tracking to estimate poses and select keyframes. The mapping stage updates the scene, followed by loop closure and optimization to ensure global consistency.}
  \label{4}
\end{figure*}

\textit{4)Scene Optimization:}
The core of 3DGS is optimizing the Gaussians to fit the scene. After rendering an image $I_{\text{render}}$, a loss between $I_{\text{render}}$ and the ground-truth image  $I_{\text{gt}}$ is computed and backpropagated to update each Gaussian’s parameters $\mu_i$,$\alpha_i$,$\Sigma_i$,$\mathbf{c}_i$.A typical loss is a weighted sum of an  $\mathcal{L}_1$ image loss and a structural similarity loss:
\begin{equation}
  \mathcal{L} = (1 - \lambda) \mathcal{L}_1 + \lambda \mathcal{L}_{\text{D-SSIM}},
\end{equation}
where $\mathcal{L}_1$ is the per-pixel L1 loss, $\mathcal{L}_{\text{D-SSIM}}$ is a multi-scale SSIM loss, and $\lambda$ weights their balance.

To manage Gaussian density, 3DGS employs adaptive splitting and merging: in over-represented areas (small positional gradients), Gaussians are split into finer ones; in under-represented regions (large gradients), new Gaussians are cloned as needed. This allows creation of Gaussians in initially missing regions while keeping dense regions well-refined, yielding an efficient representation.

\subsection{Integration of 3DGS with SLAM}
A typical 3DGS-SLAM system operates in four main stages: \textbf{initialization}, \textbf{camera tracking}, \textbf{Gaussian mapping}, and \textbf{loop closure optimization}. For example, the SplaTAM\cite{SplaTAM} system demonstrates this pipeline as follows: 

 \begin{table*}[]
\scriptsize
\caption{Summary of Performance Optimization Techniques in 3DGS-SLAM}
\label{tab_1}
\centering
\adjustbox{scale=0.96}{
\begin{NiceTabular}{l|l|l|llll|llll|ll|l|ll}
\hline
& \multicolumn{1}{c|}{} & \multicolumn{1}{c|}{}& \multicolumn{4}{c|}{Input} & \multicolumn{4}{c|}{Optimization Objective} & \multicolumn{2}{c|}{Tracking Strategy} & \multicolumn{1}{c|}{} & \multicolumn{2}{c}{Link} \\ \cline{4-13} \cline{15-16} 

\multirow{-2}{*}{Method} & \multicolumn{1}{c|}{\multirow{-2}{*}{Venue}} & \multicolumn{1}{c|}{\multirow{-2}{*}{Dataset}} & RGB  & RGBD  & IMU & Lidar & \multicolumn{1}{c}{RQ} & \multicolumn{1}{c}{TA} & \multicolumn{1}{c}{RS} & \multicolumn{1}{c|}{MC} & \multicolumn{1}{c}{F2F} & \multicolumn{1}{c|}{F2M}& \multicolumn{1}{c|}{\multirow{-2}{*}{\begin{tabular}[c]{@{}c@{}}Semantic\\ output\end{tabular}}} & \multicolumn{1}{c}{Paper} & \multicolumn{1}{c}{Code} \\ \hline

Gaussian-SLAM\cite{Gaussian-SLAM}& \multicolumn{1}{c|}{arxiv 2023} & \multicolumn{1}{c|}{R,T,S} & \multicolumn{1}{c}{} & \multicolumn{1}{c}{\checkmark} & \multicolumn{1}{c}{} & \multicolumn{1}{c|}{} & \multicolumn{1}{c}{\checkmark}  & \multicolumn{1}{c}{}  & \multicolumn{1}{c}{}  & \multicolumn{1}{c|}{}  & \multicolumn{1}{c}{} & \multicolumn{1}{c|}{\checkmark} & \multicolumn{1}{c|}{} & \href{https://arxiv.org/pdf/2312.10070}{Paper}&\href{https://github.com/VladimirYugay/gaussian-slam}{Code}\\

\rowcolor[HTML]{ECF4FF}
GS-SLAM\cite{GS-SLAM} & \multicolumn{1}{c|}{CVPR 2024}& \multicolumn{1}{c|}{R,T} & \multicolumn{1}{c}{} & \multicolumn{1}{c}{\checkmark} & \multicolumn{1}{c}{} & \multicolumn{1}{c|}{} & \multicolumn{1}{c}{}  & \multicolumn{1}{c}{}  & \multicolumn{1}{c}{\checkmark}  & \multicolumn{1}{c|}{\checkmark}  & \multicolumn{1}{c}{} & \multicolumn{1}{c|}{\checkmark} & \multicolumn{1}{c|}{} &\href{https://arxiv.org/pdf/2311.11700}{Paper}&\href{https://gs-slam.github.io/}{Code}\\

SplaTAM\cite{SplaTAM}      & \multicolumn{1}{c|}{CVPR 2024} & \multicolumn{1}{c|}{R,T,S} & \multicolumn{1}{c}{} & \multicolumn{1}{c}{\checkmark} & \multicolumn{1}{c}{} & \multicolumn{1}{c|}{} & \multicolumn{1}{c}{}  & \multicolumn{1}{c}{\checkmark}  & \multicolumn{1}{c}{}  & \multicolumn{1}{c|}{\checkmark}  & \multicolumn{1}{c}{} & \multicolumn{1}{c|}{\checkmark} & \multicolumn{1}{c|}{} &\href{https://arxiv.org/pdf/2312.02126}{Paper}&\href{https://github.com/spla-tam/SplaTAM}{Code}\\

\rowcolor[HTML]{ECF4FF}
MonoGS\cite{MonoGS}       & \multicolumn{1}{c|}{CVPR 2024} & \multicolumn{1}{c|}{R,T,E} & \multicolumn{1}{c}{\checkmark} & \multicolumn{1}{c}{\checkmark} & \multicolumn{1}{c}{} & \multicolumn{1}{c|}{} & \multicolumn{1}{c}{\checkmark}  & \multicolumn{1}{c}{}  & \multicolumn{1}{c}{}  & \multicolumn{1}{c|}{\checkmark}  & \multicolumn{1}{c}{} & \multicolumn{1}{c|}{\checkmark} & \multicolumn{1}{c|}{} &\href{https://arxiv.org/pdf/2312.06741}{Paper}&\href{https://github.com/muskie82/MonoGS}{Code}\\

Photo-SLAM\cite{Photo}   & \multicolumn{1}{c|}{CVPR 2024} & \multicolumn{1}{c|}{R,T} & \multicolumn{1}{c}{\checkmark} & \multicolumn{1}{c}{\checkmark} & \multicolumn{1}{c}{} & \multicolumn{1}{c|}{} & \multicolumn{1}{c}{\checkmark}  & \multicolumn{1}{c}{}  & \multicolumn{1}{c}{\checkmark}  & \multicolumn{1}{c|}{}  & \multicolumn{1}{c}{} & \multicolumn{1}{c|}{\checkmark} & \multicolumn{1}{c|}{} &\href{https://arxiv.org/pdf/2311.16728}{Paper}&\href{https://github.com/HuajianUP/Photo-SLAM}{Code}\\

\rowcolor[HTML]{ECF4FF}
DROID-Splat\cite{DROID-Splat}  & \multicolumn{1}{c|}{arXiv 2024} & \multicolumn{1}{c|}{R,T} & \multicolumn{1}{c}{\checkmark} & \multicolumn{1}{c}{\checkmark} & \multicolumn{1}{c}{} & \multicolumn{1}{c|}{} & \multicolumn{1}{c}{\checkmark}  & \multicolumn{1}{c}{\checkmark}  & \multicolumn{1}{c}{}  & \multicolumn{1}{c|}{}  & \multicolumn{1}{c}{} & \multicolumn{1}{c|}{\checkmark} & \multicolumn{1}{c|}{} &\href{https://arxiv.org/pdf/2411.17660}{Paper}&\href{https://github.com/ChenHoy/DROID-Splat}{Code}\\

MGS-SLAM\cite{MGS-SLAM}     & \multicolumn{1}{c|}{RA-L 2024} & \multicolumn{1}{c|}{R,T,I} & \multicolumn{1}{c}{\checkmark} & \multicolumn{1}{c}{} & \multicolumn{1}{c}{} & \multicolumn{1}{c|}{} & \multicolumn{1}{c}{\checkmark}  & \multicolumn{1}{c}{\checkmark}  & \multicolumn{1}{c}{}  & \multicolumn{1}{c|}{}  & \multicolumn{1}{c}{\checkmark} & \multicolumn{1}{c|}{} & \multicolumn{1}{c|}{} &\href{https://arxiv.org/pdf/2405.06241}{Paper}&\href{}{-}\\

\rowcolor[HTML]{ECF4FF}
GLC-SLAM\cite{GLC-SLAM}     & \multicolumn{1}{c|}{arXiv 2024} & \multicolumn{1}{c|}{R,T,S} & \multicolumn{1}{c}{} & \multicolumn{1}{c}{\checkmark} & \multicolumn{1}{c}{} & \multicolumn{1}{c|}{} & \multicolumn{1}{c}{}  & \multicolumn{1}{c}{\checkmark}  & \multicolumn{1}{c}{}  & \multicolumn{1}{c|}{}  & \multicolumn{1}{c}{} & \multicolumn{1}{c|}{\checkmark} & \multicolumn{1}{c|}{} &\href{https://arxiv.org/pdf/2409.10982}{Paper}&\href{}{-}\\

RTG-SLAM\cite{RTG-SLAM}     & \multicolumn{1}{c|}{SIGGRAPH 2024} & \multicolumn{1}{c|}{R,T,S} & \multicolumn{1}{c}{} & \multicolumn{1}{c}{\checkmark} & \multicolumn{1}{c}{} & \multicolumn{1}{c|}{} & \multicolumn{1}{c}{}  & \multicolumn{1}{c}{\checkmark}  & \multicolumn{1}{c}{\checkmark}  & \multicolumn{1}{c|}{\checkmark}  & \multicolumn{1}{c}{} & \multicolumn{1}{c|}{\checkmark} & \multicolumn{1}{c|}{} &\href{https://arxiv.org/pdf/2404.19706}{Paper}&\href{https://github.com/MisEty/RTG-SLAM}{Code}\\

\rowcolor[HTML]{ECF4FF}
GS-Loop\cite{GS-Loop}      & \multicolumn{1}{c|}{ROBIO 2024} & \multicolumn{1}{c|}{R,T} & \multicolumn{1}{c}{} & \multicolumn{1}{c}{\checkmark} & \multicolumn{1}{c}{} & \multicolumn{1}{c|}{} & \multicolumn{1}{c}{\checkmark}  & \multicolumn{1}{c}{\checkmark}  & \multicolumn{1}{c}{}  & \multicolumn{1}{c|}{}  & \multicolumn{1}{c}{} & \multicolumn{1}{c|}{\checkmark} & \multicolumn{1}{c|}{} &\href{https://ieeexplore.ieee.org/document/10907594}{Paper}&\href{}{-}\\

Mon-SLAM\cite{Mon-SLAM}     & \multicolumn{1}{c|}{arXiv 2024} & \multicolumn{1}{c|}{R,T,S,E} & \multicolumn{1}{c}{\checkmark} & \multicolumn{1}{c}{} & \multicolumn{1}{c}{} & \multicolumn{1}{c|}{} & \multicolumn{1}{c}{}  & \multicolumn{1}{c}{\checkmark}  & \multicolumn{1}{c}{}  & \multicolumn{1}{c|}{}  & \multicolumn{1}{c}{\checkmark} & \multicolumn{1}{c|}{} & \multicolumn{1}{c|}{} &\href{https://arxiv.org/pdf/2405.13748}{Paper}&\href{}{-}\\


\rowcolor[HTML]{ECF4FF}
TAMBRIDGE\cite{jiang2024tambridgebridgingframecenteredtracking}     & \multicolumn{1}{c|}{arXiv 2024} & \multicolumn{1}{c|}{T} & \multicolumn{1}{c}{} & \multicolumn{1}{c}{\checkmark} & \multicolumn{1}{c}{} & \multicolumn{1}{c|}{} & \multicolumn{1}{c}{}  & \multicolumn{1}{c}{\checkmark}  & \multicolumn{1}{c}{}  & \multicolumn{1}{c|}{}  & \multicolumn{1}{c}{} & \multicolumn{1}{c|}{\checkmark} & \multicolumn{1}{c|}{} &\href{https://arxiv.org/pdf/2405.19614}{Paper}&\href{}{-}\\

DP-SLAM\cite{DP-SLAM} & \multicolumn{1}{c|}{CEI 2024} & \multicolumn{1}{c|}{T} & \multicolumn{1}{c}{} & \multicolumn{1}{c}{\checkmark} & \multicolumn{1}{c}{} & \multicolumn{1}{c|}{} & \multicolumn{1}{c}{\checkmark}  & \multicolumn{1}{c}{}  & \multicolumn{1}{c}{}  & \multicolumn{1}{c|}{}  & \multicolumn{1}{c}{} & \multicolumn{1}{c|}{\checkmark} & \multicolumn{1}{c|}{} &\href{https://ieeexplore.ieee.org/stamp/stamp.jsp?tp=&arnumber=10871531}{Paper}&\href{}{-}\\ 

\rowcolor[HTML]{ECF4FF}
GS3LAM\cite{GS3LAM}   & \multicolumn{1}{c|}{ACM MM 2024} & \multicolumn{1}{c|}{R,S} & \multicolumn{1}{c}{} & \multicolumn{1}{c}{\checkmark} & \multicolumn{1}{c}{} & \multicolumn{1}{c|}{} & \multicolumn{1}{c}{\checkmark}  & \multicolumn{1}{c}{}  & \multicolumn{1}{c}{}  & \multicolumn{1}{c|}{}  & \multicolumn{1}{c}{} & \multicolumn{1}{c|}{\checkmark} & \multicolumn{1}{c|}{\checkmark} &\href{https://dlnext.acm.org/doi/10.1145/3664647.3680739}{Paper}&\href{https://github.com/lif314/GS3LAM}{Code}\\

RD-SLAM\cite{RD-SLAM} & \multicolumn{1}{c|}{SEP 2024} & \multicolumn{1}{c|}{R,T,S} & \multicolumn{1}{c}{} & \multicolumn{1}{c}{\checkmark} & \multicolumn{1}{c}{} & \multicolumn{1}{c|}{} & \multicolumn{1}{c}{}  & \multicolumn{1}{c}{}  & \multicolumn{1}{c}{\checkmark}  & \multicolumn{1}{c|}{}  & \multicolumn{1}{c}{} & \multicolumn{1}{c|}{\checkmark} & \multicolumn{1}{c|}{} &\href{https://www.mdpi.com/2076-3417/14/17/7767}{Paper}&\href{}{-}\\ 

\rowcolor[HTML]{ECF4FF}
GS-ICP\cite{GS-ICP}     & \multicolumn{1}{c|}{ECCV 2024} & \multicolumn{1}{c|}{R,T} & \multicolumn{1}{c}{} & \multicolumn{1}{c}{\checkmark} & \multicolumn{1}{c}{} & \multicolumn{1}{c|}{} & \multicolumn{1}{c}{}  & \multicolumn{1}{c}{}  & \multicolumn{1}{c}{\checkmark}  & \multicolumn{1}{c|}{}  & \multicolumn{1}{c}{} & \multicolumn{1}{c|}{\checkmark} & \multicolumn{1}{c|}{} &\href{https://arxiv.org/pdf/2403.12550}{Paper}&\href{https://github.com/Lab-of-AI-and-Robotics/GS_ICP_SLAM}{Code}\\ 

CG-SLAM\cite{CG-SLAM}     & \multicolumn{1}{c|}{ECCV 2024} & \multicolumn{1}{c|}{R,T,S} & \multicolumn{1}{c}{} & \multicolumn{1}{c}{\checkmark} & \multicolumn{1}{c}{} & \multicolumn{1}{c|}{} & \multicolumn{1}{c}{}  & \multicolumn{1}{c}{}  & \multicolumn{1}{c}{\checkmark}  & \multicolumn{1}{c|}{\checkmark}  & \multicolumn{1}{c}{} & \multicolumn{1}{c|}{\checkmark} & \multicolumn{1}{c|}{} &\href{https://arxiv.org/pdf/2403.16095}{Paper}&\href{https://github.com/hjr37/CG-SLAM}{Code}\\ 

\rowcolor[HTML]{ECF4FF}
SGS-SLAM\cite{SGS-SLAM}    & \multicolumn{1}{c|}{ECCV 2024} & \multicolumn{1}{c|}{R,S} & \multicolumn{1}{c}{} & \multicolumn{1}{c}{\checkmark} & \multicolumn{1}{c}{} & \multicolumn{1}{c|}{} & \multicolumn{1}{c}{\checkmark}  & \multicolumn{1}{c}{}  & \multicolumn{1}{c}{}  & \multicolumn{1}{c|}{}  & \multicolumn{1}{c}{} & \multicolumn{1}{c|}{\checkmark} & \multicolumn{1}{c|}{\checkmark} &\href{https://arxiv.org/pdf/2402.03246}{Paper}&\href{https://github.com/ShuhongLL/SGS-SLAM}{Code}\\ 

HF-SLAM\cite{HF-SLAM}      & \multicolumn{1}{c|}{IROS 2024} & \multicolumn{1}{c|}{R,T} & \multicolumn{1}{c}{} & \multicolumn{1}{c}{\checkmark} & \multicolumn{1}{c}{} & \multicolumn{1}{c|}{} & \multicolumn{1}{c}{\checkmark}  & \multicolumn{1}{c}{}  & \multicolumn{1}{c}{}  & \multicolumn{1}{c|}{}  & \multicolumn{1}{c}{} & \multicolumn{1}{c|}{\checkmark} & \multicolumn{1}{c|}{} &\href{https://arxiv.org/pdf/2403.12535}{Paper}&\href{https://github.com/ljjTYJR/HF-SLAM}{Code}\\

\rowcolor[HTML]{ECF4FF}
MM3DGS-SLAM\cite{MM3DGS-SLAM}  & \multicolumn{1}{c|}{IROS 2024} & \multicolumn{1}{c|}{T} & \multicolumn{1}{c}{\checkmark} & \multicolumn{1}{c}{\checkmark} & \multicolumn{1}{c}{\checkmark} & \multicolumn{1}{c|}{\checkmark} & \multicolumn{1}{c}{\checkmark}  & \multicolumn{1}{c}{\checkmark}  & \multicolumn{1}{c}{}  & \multicolumn{1}{c|}{}  & \multicolumn{1}{c}{} & \multicolumn{1}{c|}{\checkmark} & \multicolumn{1}{c|}{} &\href{https://arxiv.org/pdf/2404.00923}{Paper}&\href{https://github.com/VITA-Group/MM3DGS-SLAM}{Code}\\

GauSPU\cite{GauSPU}    & \multicolumn{1}{c|}{MICRO 2024} & \multicolumn{1}{c|}{R} & \multicolumn{1}{c}{} & \multicolumn{1}{c}{\checkmark} & \multicolumn{1}{c}{} & \multicolumn{1}{c|}{} & \multicolumn{1}{c}{}  & \multicolumn{1}{c}{}  & \multicolumn{1}{c}{\checkmark}  & \multicolumn{1}{c|}{}  & \multicolumn{1}{c}{} & \multicolumn{1}{c|}{\checkmark} & \multicolumn{1}{c|}{} &\href{https://ieeexplore.ieee.org/stamp/stamp.jsp?tp=&arnumber=10764611}{Paper}&\href{}{-}\\

\rowcolor[HTML]{ECF4FF}
GSFusion\cite{GSFusion}    & \multicolumn{1}{c|}{RA-L 2024}& \multicolumn{1}{c|}{R,S} & \multicolumn{1}{c}{} & \multicolumn{1}{c}{\checkmark} & \multicolumn{1}{c}{} & \multicolumn{1}{c|}{} & \multicolumn{1}{c}{}  & \multicolumn{1}{c}{}  & \multicolumn{1}{c}{}  & \multicolumn{1}{c|}{\checkmark}  & \multicolumn{1}{c}{} & \multicolumn{1}{c|}{\checkmark} & \multicolumn{1}{c|}{} &\href{https://arxiv.org/pdf/2408.12677}{Paper}&\href{https://github.com/ethz-mrl/GSFusion}{Code}\\  

MotionGS\cite{MotionGS}    & \multicolumn{1}{c|}{RCAE 2024} & \multicolumn{1}{c|}{R,T} & \multicolumn{1}{c}{\checkmark} & \multicolumn{1}{c}{\checkmark} & \multicolumn{1}{c}{} & \multicolumn{1}{c|}{} & \multicolumn{1}{c}{\checkmark}  & \multicolumn{1}{c}{}  & \multicolumn{1}{c}{}  & \multicolumn{1}{c|}{\checkmark}  & \multicolumn{1}{c}{} & \multicolumn{1}{c|}{\checkmark} & \multicolumn{1}{c|}{} &\href{https://arxiv.org/abs/2405.11129}{Paper}&\href{https://github.com/Antonio521/MotionGS}{Code}\\

\rowcolor[HTML]{ECF4FF}
Loopy-SLAM\cite{Loopy-SLAM}    & \multicolumn{1}{c|}{CVPR 2024} & \multicolumn{1}{c|}{R,T,S} & \multicolumn{1}{c}{} & \multicolumn{1}{c}{\checkmark} & \multicolumn{1}{c}{} & \multicolumn{1}{c|}{} & \multicolumn{1}{c}{}  & \multicolumn{1}{c}{\checkmark}  & \multicolumn{1}{c}{}  & \multicolumn{1}{c|}{\checkmark}  & \multicolumn{1}{c}{} & \multicolumn{1}{c|}{\checkmark} & \multicolumn{1}{c|}{} &\href{https://arxiv.org/pdf/2402.09944}{Paper}&\href{https://github.com/eriksandstroem/Loopy-SLAM}{Code}\\

FlashSLAM\cite{FlashSLAM}& \multicolumn{1}{c|}{arXiv 2024} & \multicolumn{1}{c|}{R,T} & \multicolumn{1}{c}{} & \multicolumn{1}{c}{\checkmark} & \multicolumn{1}{c}{} & \multicolumn{1}{c|}{} & \multicolumn{1}{c}{\checkmark}  & \multicolumn{1}{c}{}  & \multicolumn{1}{c}{}  & \multicolumn{1}{c|}{\checkmark}  & \multicolumn{1}{c}{\checkmark} & \multicolumn{1}{c|}{} & \multicolumn{1}{c|}{} &\href{https://arxiv.org/pdf/2412.00682}{Paper}&\href{}{-}\\ 

\rowcolor[HTML]{ECF4FF}
NEDS-SLAM\cite{NEDS-SLAM}& \multicolumn{1}{c|}{RA-L 2024} & \multicolumn{1}{c|}{R,S} & \multicolumn{1}{c}{} & \multicolumn{1}{c}{\checkmark} & \multicolumn{1}{c}{} & \multicolumn{1}{c|}{} & \multicolumn{1}{c}{}  & \multicolumn{1}{c}{}  & \multicolumn{1}{c}{\checkmark}  & \multicolumn{1}{c|}{\checkmark}  & \multicolumn{1}{c}{} & \multicolumn{1}{c|}{\checkmark} & \multicolumn{1}{c|}{\checkmark} &\href{https://arxiv.org/pdf/2403.11679}{Paper}&\href{}{}\\ 

LIV-GaussMap\cite{LIV-GaussMap}& \multicolumn{1}{c|}{RA-L 2024} & \multicolumn{1}{c|}{F} & \multicolumn{1}{c}{\checkmark} & \multicolumn{1}{c}{} & \multicolumn{1}{c}{\checkmark} & \multicolumn{1}{c|}{\checkmark} & \multicolumn{1}{c}{\checkmark}  & \multicolumn{1}{c}{}  & \multicolumn{1}{c}{}  & \multicolumn{1}{c|}{}  & \multicolumn{1}{c}{} & \multicolumn{1}{c|}{\checkmark} & \multicolumn{1}{c|}{} &\href{https://arxiv.org/pdf/2401.14857}{Paper}&\href{https://github.com/sheng00125/LIV-GaussMap}{Code}\\

\rowcolor[HTML]{ECF4FF}
VINGS-Mono\cite{VINGS-Mono} &\multicolumn{1}{c|}{TRO 2025} & \multicolumn{1}{c|}{K,W} & \multicolumn{1}{c}{\checkmark} & \multicolumn{1}{c}{} & \multicolumn{1}{c}{\checkmark} & \multicolumn{1}{c|}{} & \multicolumn{1}{c}{\checkmark}  & \multicolumn{1}{c}{}  & \multicolumn{1}{c}{}  & \multicolumn{1}{c|}{}  & \multicolumn{1}{c}{} & \multicolumn{1}{c|}{\checkmark} & \multicolumn{1}{c|}{} &\href{https://arxiv.org/pdf/2501.08286}{Paper}&\href{https://github.com/Fudan-MAGIC-Lab/VINGS-Mono}{Code}\\

Gaussian-LIC\cite{Gaussian-LIC}& \multicolumn{1}{c|}{ICRA 2025} & \multicolumn{1}{c|}{F} & \multicolumn{1}{c}{\checkmark} & \multicolumn{1}{c}{} & \multicolumn{1}{c}{\checkmark} & \multicolumn{1}{c|}{\checkmark} & \multicolumn{1}{c}{\checkmark}  & \multicolumn{1}{c}{}  & \multicolumn{1}{c}{}  & \multicolumn{1}{c|}{}  & \multicolumn{1}{c}{} & \multicolumn{1}{c|}{\checkmark} & \multicolumn{1}{c|}{} &\href{https://arxiv.org/pdf/2404.06926}{Paper}&\href{https://github.com/APRIL-ZJU/Gaussian-LIC}{Code}\\

\rowcolor[HTML]{ECF4FF}
GI-SLAM\cite{GI-SLAM} & \multicolumn{1}{c|}{arXiv 2025}& \multicolumn{1}{c|}{T,E} & \multicolumn{1}{c}{\checkmark} & \multicolumn{1}{c}{\checkmark} & \multicolumn{1}{c}{\checkmark} & \multicolumn{1}{c|}{} & \multicolumn{1}{c}{}  & \multicolumn{1}{c}{\checkmark}  & \multicolumn{1}{c}{}  & \multicolumn{1}{c|}{}  & \multicolumn{1}{c}{} & \multicolumn{1}{c|}{\checkmark} & \multicolumn{1}{c|}{} &\href{https://arxiv.org/pdf/2503.18275}{Paper}&\href{}{-}\\

NGM-SLAM\cite{NGM-SLAM}     & \multicolumn{1}{c|}{arXiv 2025}& \multicolumn{1}{c|}{R,T,S,E} & \multicolumn{1}{c}{\checkmark} & \multicolumn{1}{c}{\checkmark} & \multicolumn{1}{c}{} & \multicolumn{1}{c|}{} & \multicolumn{1}{c}{\checkmark}  & \multicolumn{1}{c}{\checkmark}  & \multicolumn{1}{c}{}  & \multicolumn{1}{c|}{\checkmark}  & \multicolumn{1}{c}{} & \multicolumn{1}{c|}{\checkmark} & \multicolumn{1}{c|}{} &\href{https://arxiv.org/pdf/2405.05702}{Paper}&\href{}{-}\\

\rowcolor[HTML]{ECF4FF}
DenseSplat\cite{DenseSplat}   & \multicolumn{1}{c|}{arXiv 2025}& \multicolumn{1}{c|}{R,T,S} & \multicolumn{1}{c}{} & \multicolumn{1}{c}{\checkmark} & \multicolumn{1}{c}{} & \multicolumn{1}{c|}{} & \multicolumn{1}{c}{\checkmark}  & \multicolumn{1}{c}{\checkmark}  & \multicolumn{1}{c}{}  & \multicolumn{1}{c|}{\checkmark}  & \multicolumn{1}{c}{} & \multicolumn{1}{c|}{\checkmark} & \multicolumn{1}{c|}{} &\href{https://arxiv.org/pdf/2502.09111}{Paper}&\href{}{-}\\

GSFF-SLAM\cite{GSFF-SLAM}& \multicolumn{1}{c|}{arXiv 2025}& \multicolumn{1}{c|}{R,T,S} & \multicolumn{1}{c}{} & \multicolumn{1}{c}{\checkmark} & \multicolumn{1}{c}{} & \multicolumn{1}{c|}{} & \multicolumn{1}{c}{\checkmark}  & \multicolumn{1}{c}{}  & \multicolumn{1}{c}{}  & \multicolumn{1}{c|}{}  & \multicolumn{1}{c}{} & \multicolumn{1}{c|}{\checkmark} & \multicolumn{1}{c|}{\checkmark} &\href{https://arxiv.org/pdf/2504.19409}{Paper}&\href{}{-}\\

\rowcolor[HTML]{ECF4FF}
MemGS\cite{MemGS}& \multicolumn{1}{c|}{arXiv 2025}& \multicolumn{1}{c|}{R,T} & \multicolumn{1}{c}{\checkmark} & \multicolumn{1}{c}{\checkmark} & \multicolumn{1}{c}{} & \multicolumn{1}{c|}{} & \multicolumn{1}{c}{}  & \multicolumn{1}{c}{}  & \multicolumn{1}{c}{\checkmark}  & \multicolumn{1}{c|}{\checkmark}  & \multicolumn{1}{c}{} & \multicolumn{1}{c|}{\checkmark} & \multicolumn{1}{c|}{} &\href{https://arxiv.org/pdf/2509.13536}
{Paper}&\href{https://github.com/NAIL-HNU/MemGS_SLAM}{Code}\\

G2S-ICP\cite{G2S-ICP}& \multicolumn{1}{c|}{arXiv 2025}& \multicolumn{1}{c|}{R,T} & \multicolumn{1}{c}{} & \multicolumn{1}{c}{\checkmark} & \multicolumn{1}{c}{} & \multicolumn{1}{c|}{} & \multicolumn{1}{c}{}  & \multicolumn{1}{c}{}  & \multicolumn{1}{c}{\checkmark}  & \multicolumn{1}{c|}{}  & \multicolumn{1}{c}{} & \multicolumn{1}{c|}{\checkmark} & \multicolumn{1}{c|}{} &\href{https://arxiv.org/pdf/2507.18344}{Paper}&\href{}{-}\\

\rowcolor[HTML]{ECF4FF}
Constrained\cite{Constrained}  & \multicolumn{1}{c|}{TAI 2025}& \multicolumn{1}{c|}{R,T,S} & \multicolumn{1}{c}{} & \multicolumn{1}{c}{\checkmark} & \multicolumn{1}{c}{} & \multicolumn{1}{c|}{} & \multicolumn{1}{c}{\checkmark}  & \multicolumn{1}{c}{\checkmark}  & \multicolumn{1}{c}{}  & \multicolumn{1}{c|}{}  & \multicolumn{1}{c}{} & \multicolumn{1}{c|}{\checkmark} & \multicolumn{1}{c|}{} &\href{https://ieeexplore.ieee.org/stamp/stamp.jsp?tp=&arnumber=11060934}{Paper}&\href{}{-}\\

MG-SLAM\cite{MG-SLAM}      & \multicolumn{1}{c|}{T-ASE 2025}& \multicolumn{1}{c|}{R,S} & \multicolumn{1}{c}{} & \multicolumn{1}{c}{\checkmark} & \multicolumn{1}{c}{} & \multicolumn{1}{c|}{} & \multicolumn{1}{c}{\checkmark}  & \multicolumn{1}{c}{\checkmark}  & \multicolumn{1}{c}{}  & \multicolumn{1}{c|}{}  & \multicolumn{1}{c}{\checkmark} & \multicolumn{1}{c|}{} & \multicolumn{1}{c|}{\checkmark} &\href{https://arxiv.org/pdf/2405.20031v1}{Paper}&\href{}{}\\

\rowcolor[HTML]{ECF4FF}
Splat-SLAM\cite{Splat-SLAM}   & \multicolumn{1}{c|}{CVPRW 2025} & \multicolumn{1}{c|}{R,T,S} & \multicolumn{1}{c}{\checkmark} & \multicolumn{1}{c}{} & \multicolumn{1}{c}{} & \multicolumn{1}{c|}{} & \multicolumn{1}{c}{\checkmark}  & \multicolumn{1}{c}{\checkmark}  & \multicolumn{1}{c}{}  & \multicolumn{1}{c|}{}  & \multicolumn{1}{c}{\checkmark} & \multicolumn{1}{c|}{} & \multicolumn{1}{c|}{} &\href{https://arxiv.org/pdf/2405.16544}{Paper}&\href{https://github.com/google-research/Splat-SLAM}{Code}\\

MVS-GS\cite{MVS-GS}       & \multicolumn{1}{c|}{Access 2025}& \multicolumn{1}{c|}{R,T} & \multicolumn{1}{c}{\checkmark} & \multicolumn{1}{c}{} & \multicolumn{1}{c}{} & \multicolumn{1}{c|}{} & \multicolumn{1}{c}{\checkmark}  & \multicolumn{1}{c}{}  & \multicolumn{1}{c}{}  & \multicolumn{1}{c|}{}  & \multicolumn{1}{c}{} & \multicolumn{1}{c|}{\checkmark} & \multicolumn{1}{c|}{} &\href{https://arxiv.org/pdf/2412.19130}{Paper}&\href{}{-}\\

\rowcolor[HTML]{ECF4FF}
SplatMAP\cite{SplatMAP}     & \multicolumn{1}{c|}{PACMCGIT 2025} & \multicolumn{1}{c|}{R,T} & \multicolumn{1}{c}{\checkmark} & \multicolumn{1}{c}{} & \multicolumn{1}{c}{} & \multicolumn{1}{c|}{} & \multicolumn{1}{c}{\checkmark}  & \multicolumn{1}{c}{\checkmark}  & \multicolumn{1}{c}{}  & \multicolumn{1}{c|}{}  & \multicolumn{1}{c}{} & \multicolumn{1}{c|}{\checkmark} & \multicolumn{1}{c|}{} &\href{https://arxiv.org/pdf/2501.07015}{Paper}&\href{}{-}\\

OGS-SLAM\cite{OGS-SLAM}& \multicolumn{1}{c|}{AAMAS 2025} & \multicolumn{1}{c|}{R,T} & \multicolumn{1}{c}{} & \multicolumn{1}{c}{\checkmark} & \multicolumn{1}{c}{} & \multicolumn{1}{c|}{} & \multicolumn{1}{c}{}  & \multicolumn{1}{c}{\checkmark}  & \multicolumn{1}{c}{}  & \multicolumn{1}{c|}{}  & \multicolumn{1}{c}{} & \multicolumn{1}{c|}{\checkmark} & \multicolumn{1}{c|}{\checkmark} &\href{https://aamas.csc.liv.ac.uk/Proceedings/aamas2025/pdfs/p1300.pdf}{Paper}&\href{}{-}\\

\rowcolor[HTML]{ECF4FF}
LoopSplat\cite{LoopSplat}     & \multicolumn{1}{c|}{3DV 2025}& \multicolumn{1}{c|}{R,T,S} & \multicolumn{1}{c}{} & \multicolumn{1}{c}{\checkmark} & \multicolumn{1}{c}{} & \multicolumn{1}{c|}{} & \multicolumn{1}{c}{}  & \multicolumn{1}{c}{\checkmark}  & \multicolumn{1}{c}{}  & \multicolumn{1}{c|}{}  & \multicolumn{1}{c}{} & \multicolumn{1}{c|}{\checkmark} & \multicolumn{1}{c|}{} &\href{https://arxiv.org/pdf/2408.10154}{Paper}&\href{https://github.com/GradientSpaces/loopsplat}{Code}\\

Scaffold-SLAM\cite{Scaffold-SLAM}& \multicolumn{1}{c|}{arXiv 2025} & \multicolumn{1}{c|}{R,T,E} & \multicolumn{1}{c}{\checkmark} & \multicolumn{1}{c}{\checkmark} & \multicolumn{1}{c}{} & \multicolumn{1}{c|}{} & \multicolumn{1}{c}{\checkmark}  & \multicolumn{1}{c}{}  & \multicolumn{1}{c}{}  & \multicolumn{1}{c|}{}  & \multicolumn{1}{c}{} & \multicolumn{1}{c|}{\checkmark} & \multicolumn{1}{c|}{} &\href{https://arxiv.org/pdf/2501.05242v1}{Paper}&\href{}{-}\\

\rowcolor[HTML]{ECF4FF}
FIGS-SLAM\cite{FIGS-SLAM} & \multicolumn{1}{c|}{ESWA 2025} & \multicolumn{1}{c|}{R,T} & \multicolumn{1}{c}{} & \multicolumn{1}{c}{\checkmark} & \multicolumn{1}{c}{} & \multicolumn{1}{c|}{} & \multicolumn{1}{c}{\checkmark}  & \multicolumn{1}{c}{}  & \multicolumn{1}{c}{}  & \multicolumn{1}{c|}{}  & \multicolumn{1}{c}{} & \multicolumn{1}{c|}{\checkmark} & \multicolumn{1}{c|}{} &\href{https://www.sciencedirect.com/science/article/abs/pii/S0957417425023814}{Paper}&\href{}{-}\\

LVI-GS\cite{LVI-GS}    & \multicolumn{1}{c|}{T-IM 2025} & \multicolumn{1}{c|}{F} & \multicolumn{1}{c}{\checkmark} & \multicolumn{1}{c}{} & \multicolumn{1}{c}{\checkmark} & \multicolumn{1}{c|}{\checkmark} & \multicolumn{1}{c}{\checkmark}  & \multicolumn{1}{c}{}  & \multicolumn{1}{c}{}  & \multicolumn{1}{c|}{}  & \multicolumn{1}{c}{} & \multicolumn{1}{c|}{\checkmark} & \multicolumn{1}{c|}{} &\href{https://arxiv.org/pdf/2411.02703}{Paper}&\href{https://kwanwaipang.github.io/LVI-GS/}{Code}\\

\rowcolor[HTML]{ECF4FF}
FT-SLAM\cite{FT-SLAM} & \multicolumn{1}{c|}{ICARA 2025}& \multicolumn{1}{c|}{T,E} & \multicolumn{1}{c}{} & \multicolumn{1}{c}{\checkmark} & \multicolumn{1}{c}{} & \multicolumn{1}{c|}{} & \multicolumn{1}{c}{}  & \multicolumn{1}{c}{\checkmark}  & \multicolumn{1}{c}{}  & \multicolumn{1}{c|}{}  & \multicolumn{1}{c}{} & \multicolumn{1}{c|}{\checkmark} & \multicolumn{1}{c|}{} &\href{https://ieeexplore.ieee.org/stamp/stamp.jsp?tp=&arnumber=10977682}{Paper}&\href{}{-}\\

MAGiC-SLAM\cite{MAGiC-SLAM}    & \multicolumn{1}{c|}{CVPR 2025} & \multicolumn{1}{c|}{MR} & \multicolumn{1}{c}{} & \multicolumn{1}{c}{\checkmark} & \multicolumn{1}{c}{} & \multicolumn{1}{c|}{} & \multicolumn{1}{c}{\checkmark}  & \multicolumn{1}{c}{}  & \multicolumn{1}{c}{}  & \multicolumn{1}{c|}{}  & \multicolumn{1}{c}{\checkmark} & \multicolumn{1}{c|}{} & \multicolumn{1}{c|}{} &\href{https://arxiv.org/pdf/2411.16785}{Paper}&\href{https://github.com/VladimirYugay/MAGiC-SLAM}{Code}\\

\rowcolor[HTML]{ECF4FF}
GRAND-SLAM\cite{GRAND-SLAM}    & \multicolumn{1}{c|}{RA-L 2025} & \multicolumn{1}{c|}{MR} & \multicolumn{1}{c}{} & \multicolumn{1}{c}{\checkmark} & \multicolumn{1}{c}{} & \multicolumn{1}{c|}{} & \multicolumn{1}{c}{\checkmark}  & \multicolumn{1}{c}{}  & \multicolumn{1}{c}{}  & \multicolumn{1}{c|}{}  & \multicolumn{1}{c}{\checkmark} & \multicolumn{1}{c|}{} & \multicolumn{1}{c|}{} &\href{https://arxiv.org/pdf/2506.18885}{Paper}&\href{}{-}\\

CompactGS\cite{CompactGS}& \multicolumn{1}{c|}{SENS J 2025} & \multicolumn{1}{c|}{R,T} & \multicolumn{1}{c}{} & \multicolumn{1}{c}{\checkmark} & \multicolumn{1}{c}{} & \multicolumn{1}{c|}{} & \multicolumn{1}{c}{}  & \multicolumn{1}{c}{}  & \multicolumn{1}{c}{}  & \multicolumn{1}{c|}{\checkmark}  & \multicolumn{1}{c}{} & \multicolumn{1}{c|}{\checkmark} & \multicolumn{1}{c|}{} &\href{https://ieeexplore.ieee.org/stamp/stamp.jsp?tp=&arnumber=11006971}{Paper}&\href{}{-}\\ 

\rowcolor[HTML]{ECF4FF}
DSOSplat\cite{DSOSplat}& \multicolumn{1}{c|}{SENS J 2025} & \multicolumn{1}{c|}{R,S} & \multicolumn{1}{c}{\checkmark} & \multicolumn{1}{c}{} & \multicolumn{1}{c}{} & \multicolumn{1}{c|}{} & \multicolumn{1}{c}{}  & \multicolumn{1}{c}{\checkmark}  & \multicolumn{1}{c}{}  & \multicolumn{1}{c|}{}  & \multicolumn{1}{c}{\checkmark} & \multicolumn{1}{c|}{} & \multicolumn{1}{c|}{} &\href{https://ieeexplore.ieee.org/stamp/stamp.jsp?tp=&arnumber=10994228}{Paper}&\href{}{-}\\ 

OpenGS-SLAM\cite{OpenGS-SLAM}          & \multicolumn{1}{c|}{ICRA 2025}& \multicolumn{1}{c|}{W} & \multicolumn{1}{c}{\checkmark} & \multicolumn{1}{c}{} & \multicolumn{1}{c}{} & \multicolumn{1}{c|}{} & \multicolumn{1}{c}{\checkmark}  & \multicolumn{1}{c}{}  & \multicolumn{1}{c}{}  & \multicolumn{1}{c|}{}  & \multicolumn{1}{c}{} & \multicolumn{1}{c|}{\checkmark} & \multicolumn{1}{c|}{} &\href{https://arxiv.org/pdf/2502.15633}{Paper}&\href{https://github.com/3DAgentWorld/OpenGS-SLAM}{Code}\\

\rowcolor[HTML]{ECF4FF}
MGSO \cite{MGSO}         & \multicolumn{1}{c|}{ICRA 2025} & \multicolumn{1}{c|}{R,T,E} & \multicolumn{1}{c}{\checkmark} & \multicolumn{1}{c}{} & \multicolumn{1}{c}{} & \multicolumn{1}{c|}{} & \multicolumn{1}{c}{}  & \multicolumn{1}{c}{}  & \multicolumn{1}{c}{\checkmark}  & \multicolumn{1}{c|}{\checkmark}  & \multicolumn{1}{c}{\checkmark} & \multicolumn{1}{c|}{} & \multicolumn{1}{c|}{} &\href{https://arxiv.org/pdf/2409.13055}{Paper}&\href{}{-}\\

MonoGS++\cite{MonoGS++}    & \multicolumn{1}{c|}{arXiv 2025} & \multicolumn{1}{c|}{R,T} & \multicolumn{1}{c}{\checkmark} & \multicolumn{1}{c}{} & \multicolumn{1}{c}{} & \multicolumn{1}{c|}{} & \multicolumn{1}{c}{}  & \multicolumn{1}{c}{}  & \multicolumn{1}{c}{\checkmark}  & \multicolumn{1}{c|}{}  & \multicolumn{1}{c}{\checkmark} & \multicolumn{1}{c|}{} & \multicolumn{1}{c|}{} &\href{https://arxiv.org/pdf/2504.02437}{Paper}&\href{}{-}\\

\rowcolor[HTML]{ECF4FF}
RGBDS-SLAM\cite{RGBDS-SLAM}    & \multicolumn{1}{c|}{RA-L 2025} & \multicolumn{1}{c|}{R,S} & \multicolumn{1}{c}{} & \multicolumn{1}{c}{\checkmark} & \multicolumn{1}{c}{} & \multicolumn{1}{c|}{} & \multicolumn{1}{c}{\checkmark}  & \multicolumn{1}{c}{}  & \multicolumn{1}{c}{}  & \multicolumn{1}{c|}{}  & \multicolumn{1}{c}{} & \multicolumn{1}{c|}{\checkmark} & \multicolumn{1}{c|}{\checkmark} &\href{https://arxiv.org/pdf/2412.01217}{Paper}&\href{https://github.com/zhenzhongcao/RGBDS-SLAM}{Code}\\

HI-SLAM2\cite{HI-SLAM2}      & \multicolumn{1}{c|}{T-RO 2025}& \multicolumn{1}{c|}{R,S,W} & \multicolumn{1}{c}{\checkmark} & \multicolumn{1}{c}{} & \multicolumn{1}{c}{} & \multicolumn{1}{c|}{} & \multicolumn{1}{c}{}  & \multicolumn{1}{c}{\checkmark}  & \multicolumn{1}{c}{}  & \multicolumn{1}{c|}{}  & \multicolumn{1}{c}{\checkmark} & \multicolumn{1}{c|}{} & \multicolumn{1}{c|}{} &\href{https://ieeexplore.ieee.org/stamp/stamp.jsp?tp=&arnumber=11219329}{Paper}&\href{https://hi-slam2.github.io/}{Code}\\

\rowcolor[HTML]{ECF4FF}
GS-LIVO\cite{GS-LIVO}& \multicolumn{1}{c|}{T-RO 2025} & \multicolumn{1}{c|}{F} & \multicolumn{1}{c}{\checkmark} & \multicolumn{1}{c}{} & \multicolumn{1}{c}{\checkmark} & \multicolumn{1}{c|}{\checkmark} & \multicolumn{1}{c}{\checkmark}  & \multicolumn{1}{c}{}  & \multicolumn{1}{c}{}  & \multicolumn{1}{c|}{}  & \multicolumn{1}{c}{} & \multicolumn{1}{c|}{\checkmark} & \multicolumn{1}{c|}{} &\href{https://arxiv.org/pdf/2501.08672}{Paper}&\href{https://github.com/HKUST-Aerial-Robotics/GS-LIVO}{Code}\\

FGO-SLAM\cite{FGO-SLAM}          & \multicolumn{1}{c|}{ICRA 2025} & \multicolumn{1}{c|}{R,T,S} & \multicolumn{1}{c}{\checkmark} & \multicolumn{1}{c}{\checkmark} & \multicolumn{1}{c}{} & \multicolumn{1}{c|}{} & \multicolumn{1}{c}{\checkmark}  & \multicolumn{1}{c}{\checkmark}  & \multicolumn{1}{c}{}  & \multicolumn{1}{c|}{}  & \multicolumn{1}{c}{\checkmark} & \multicolumn{1}{c|}{} & \multicolumn{1}{c|}{} &\href{https://ieeexplore.ieee.org/stamp/stamp.jsp?tp=&arnumber=11128403&tag=1}{Paper}&\href{}{-}\\

\rowcolor[HTML]{ECF4FF}
GS4\cite{GS4}    & \multicolumn{1}{c|}{arXiv 2025} & \multicolumn{1}{c|}{T,S} & \multicolumn{1}{c}{} & \multicolumn{1}{c}{\checkmark} & \multicolumn{1}{c}{} & \multicolumn{1}{c|}{} & \multicolumn{1}{c}{\checkmark}  & \multicolumn{1}{c}{}  & \multicolumn{1}{c}{}  & \multicolumn{1}{c|}{}  & \multicolumn{1}{c}{\checkmark} & \multicolumn{1}{c|}{} & \multicolumn{1}{c|}{\checkmark} &\href{https://arxiv.org/pdf/2506.06517}{Paper}&\href{https://mingqij.github.io/projects/gs4/}{Code}\\

G2S-SLAM\cite{G2S-SLAM}& \multicolumn{1}{c|}{CCC 2025} & \multicolumn{1}{c|}{R,T} & \multicolumn{1}{c}{} & \multicolumn{1}{c}{\checkmark} & \multicolumn{1}{c}{} & \multicolumn{1}{c|}{} & \multicolumn{1}{c}{}  & \multicolumn{1}{c}{}  & \multicolumn{1}{c}{\checkmark}  & \multicolumn{1}{c|}{}  & \multicolumn{1}{c}{} & \multicolumn{1}{c|}{\checkmark} & \multicolumn{1}{c|}{} &\href{https://ieeexplore.ieee.org/stamp/stamp.jsp?tp=&arnumber=11179229}{Paper}&\href{}{-}\\

\rowcolor[HTML]{ECF4FF}
MSGS-SLAM\cite{MSGS-SLAM}& \multicolumn{1}{c|}{Symmetry 2025} & \multicolumn{1}{c|}{R,S} & \multicolumn{1}{c}{\checkmark} & \multicolumn{1}{c}{} & \multicolumn{1}{c}{} & \multicolumn{1}{c|}{} & \multicolumn{1}{c}{}  & \multicolumn{1}{c}{\checkmark}  & \multicolumn{1}{c}{}  & \multicolumn{1}{c|}{}  & \multicolumn{1}{c}{\checkmark} & \multicolumn{1}{c|}{} & \multicolumn{1}{c|}{\checkmark} &\href{https://www.mdpi.com/2073-8994/17/9/1576}{Paper}&\href{}{-}\\

SAGA-SLAM\cite{SAGA-SLAM}    & \multicolumn{1}{c|}{RA-L 2025} & \multicolumn{1}{c|}{R,T,K} & \multicolumn{1}{c}{} & \multicolumn{1}{c}{\checkmark} & \multicolumn{1}{c}{} & \multicolumn{1}{c|}{} & \multicolumn{1}{c}{}  & \multicolumn{1}{c}{}  & \multicolumn{1}{c}{\checkmark}  & \multicolumn{1}{c|}{}  & \multicolumn{1}{c}{} & \multicolumn{1}{c|}{\checkmark} & \multicolumn{1}{c|}{} &\href{https://ieeexplore.ieee.org/stamp/stamp.jsp?tp=&arnumber=11067946}{Paper}&\href{}{-}\\ 

\rowcolor[HTML]{ECF4FF}
GSORB-SLAM\cite{GSORB-SLAM}    & \multicolumn{1}{c|}{RA-L 2025} & \multicolumn{1}{c|}{R,T,S} & \multicolumn{1}{c}{} & \multicolumn{1}{c}{\checkmark} & \multicolumn{1}{c}{} & \multicolumn{1}{c|}{} & \multicolumn{1}{c}{}  & \multicolumn{1}{c}{\checkmark}  & \multicolumn{1}{c}{}  & \multicolumn{1}{c|}{}  & \multicolumn{1}{c}{} & \multicolumn{1}{c|}{\checkmark} & \multicolumn{1}{c|}{} &\href{https://ieeexplore.ieee.org/document/11091447}{Paper}&\href{https://github.com/Aczheng-cai/GSORB-SLAM}{Code}\\

CaRtGS\cite{CaRtGS}& \multicolumn{1}{c|}{RA-L 2025} & \multicolumn{1}{c|}{R,T} & \multicolumn{1}{c}{\checkmark} & \multicolumn{1}{c}{\checkmark} & \multicolumn{1}{c}{} & \multicolumn{1}{c|}{} & \multicolumn{1}{c}{}  & \multicolumn{1}{c}{}  & \multicolumn{1}{c}{\checkmark}  & \multicolumn{1}{c|}{}  & \multicolumn{1}{c}{\checkmark} & \multicolumn{1}{c|}{} & \multicolumn{1}{c|}{} &\href{https://ieeexplore.ieee.org/stamp/stamp.jsp?tp=&arnumber=10900401&utm_source=clarivate&getft_integrator=clarivate}{Paper}&\href{https://github.com/DapengFeng/cartgs}{Code}\\ 

\rowcolor[HTML]{ECF4FF}
SGR-SLAM\cite{SGR-SLAM}    & \multicolumn{1}{c|}{RA-L 2025} & \multicolumn{1}{c|}{R,T,E} & \multicolumn{1}{c}{\checkmark} & \multicolumn{1}{c}{} & \multicolumn{1}{c}{} & \multicolumn{1}{c|}{} & \multicolumn{1}{c}{}  & \multicolumn{1}{c}{}  & \multicolumn{1}{c}{\checkmark}  & \multicolumn{1}{c|}{}  & \multicolumn{1}{c}{\checkmark} & \multicolumn{1}{c|}{} & \multicolumn{1}{c|}{} &\href{https://ieeexplore.ieee.org/stamp/stamp.jsp?tp=&arnumber=11052667}{Paper}&\href{}{-}\\

KAIST-SLAM\cite{kaist}& \multicolumn{1}{c|}{ISCAS 2025} & \multicolumn{1}{c|}{R} & \multicolumn{1}{c}{} & \multicolumn{1}{c}{\checkmark} & \multicolumn{1}{c}{} & \multicolumn{1}{c|}{} & \multicolumn{1}{c}{}  & \multicolumn{1}{c}{}  & \multicolumn{1}{c}{\checkmark}  & \multicolumn{1}{c|}{}  & \multicolumn{1}{c}{} & \multicolumn{1}{c|}{\checkmark} & \multicolumn{1}{c|}{} &\href{https://ieeexplore.ieee.org/stamp/stamp.jsp?tp=&arnumber=11043574}{Paper}&\href{}{-}\\ 

\rowcolor[HTML]{ECF4FF}
VPGS-SLAM\cite{VPGS-SLAM}& \multicolumn{1}{c|}{arXiv 2025} & \multicolumn{1}{c|}{R,K} & \multicolumn{1}{c}{} & \multicolumn{1}{c}{\checkmark} & \multicolumn{1}{c}{} & \multicolumn{1}{c|}{\checkmark} & \multicolumn{1}{c}{}  & \multicolumn{1}{c}{}  & \multicolumn{1}{c}{}  & \multicolumn{1}{c|}{\checkmark}  & \multicolumn{1}{c}{} & \multicolumn{1}{c|}{\checkmark} & \multicolumn{1}{c|}{} &\href{https://arxiv.org/pdf/2505.18992}{Paper}&\href{https://github.com/dtc111111/vpgs-slam}{Code}\\

2DGS-SLAM\cite{2DGS-SLAM} & \multicolumn{1}{c|}{arXiv 2025} & \multicolumn{1}{c|}{R,T,S} & \multicolumn{1}{c}{} & \multicolumn{1}{c}{\checkmark} & \multicolumn{1}{c}{} & \multicolumn{1}{c|}{} & \multicolumn{1}{c}{\checkmark}  & \multicolumn{1}{c}{}  & \multicolumn{1}{c}{}  & \multicolumn{1}{c|}{\checkmark}  & \multicolumn{1}{c}{} & \multicolumn{1}{c|}{\checkmark} & \multicolumn{1}{c|}{} &\href{https://arxiv.org/pdf/2506.00970}{Paper}&\href{https://github.com/PRBonn/2DGS-SLAM}{Code}\\ 

\rowcolor[HTML]{ECF4FF}
S3LAM\cite{S3LAM} & \multicolumn{1}{c|}{arXiv 2025} & \multicolumn{1}{c|}{R,T,S} & \multicolumn{1}{c}{} & \multicolumn{1}{c}{\checkmark} & \multicolumn{1}{c}{} & \multicolumn{1}{c|}{} & \multicolumn{1}{c}{}  & \multicolumn{1}{c}{\checkmark}  & \multicolumn{1}{c}{}  & \multicolumn{1}{c|}{\checkmark}  & \multicolumn{1}{c}{} & \multicolumn{1}{c|}{\checkmark} & \multicolumn{1}{c|}{} &\href{https://arxiv.org/pdf/2507.20854}{Paper}&\href{}{-}\\

OmniMap\cite{OmniMap}& \multicolumn{1}{c|}{TRO 2025} & \multicolumn{1}{c|}{R,S} & \multicolumn{1}{c}{} & \multicolumn{1}{c}{\checkmark} & \multicolumn{1}{c}{} & \multicolumn{1}{c|}{} & \multicolumn{1}{c}{}  & \multicolumn{1}{c}{}  & \multicolumn{1}{c}{}  & \multicolumn{1}{c|}{\checkmark}  & \multicolumn{1}{c}{} & \multicolumn{1}{c|}{\checkmark} & \multicolumn{1}{c|}{\checkmark} &\href{https://ieeexplore.ieee.org/stamp/stamp.jsp?tp=&arnumber=11203277}{Paper}&\href{https://github.com/BIT-DYN/omnimap}{Code}\\

\rowcolor[HTML]{ECF4FF}
CGS-SLAM\cite{CGS-SLAM}& \multicolumn{1}{c|}{IROS 2025} & \multicolumn{1}{c|}{R,S} & \multicolumn{1}{c}{} & \multicolumn{1}{c}{\checkmark} & \multicolumn{1}{c}{} & \multicolumn{1}{c|}{} & \multicolumn{1}{c}{}  & \multicolumn{1}{c}{}  & \multicolumn{1}{c}{}  & \multicolumn{1}{c|}{\checkmark}  & \multicolumn{1}{c}{} & \multicolumn{1}{c|}{\checkmark} & \multicolumn{1}{c|}{} &\href{https://www.semanticscholar.org/reader/287a070fd686c4a0432b1d5415fbc2e991025ff1}{Paper}&\href{}{-}\\

SemGauss-SLAM\cite{SemGauss-SLAM}     & \multicolumn{1}{c|}{IROS 2025}& \multicolumn{1}{c|}{R,S} & \multicolumn{1}{c}{} & \multicolumn{1}{c}{\checkmark} & \multicolumn{1}{c}{} & \multicolumn{1}{c|}{} & \multicolumn{1}{c}{\checkmark}  & \multicolumn{1}{c}{}  & \multicolumn{1}{c}{}  & \multicolumn{1}{c|}{}  & \multicolumn{1}{c}{} & \multicolumn{1}{c|}{\checkmark} & \multicolumn{1}{c|}{\checkmark} &\href{https://arxiv.org/pdf/2403.07494}{Paper}&\href{https://github.com/IRMVLab/SemGauss-SLAM}{Code}\\

\rowcolor[HTML]{ECF4FF}
GPS-SLAM\cite{GPS-SLAM}& \multicolumn{1}{c|}{CVM 2025} & \multicolumn{1}{c|}{R,T,S} & \multicolumn{1}{c}{} & \multicolumn{1}{c}{\checkmark} & \multicolumn{1}{c}{} & \multicolumn{1}{c|}{} & \multicolumn{1}{c}{}  & \multicolumn{1}{c}{}  & \multicolumn{1}{c}{\checkmark}  & \multicolumn{1}{c|}{\checkmark}  & \multicolumn{1}{c}{} & \multicolumn{1}{c|}{\checkmark} & \multicolumn{1}{c|}{} &\href{https://ieeexplore.ieee.org/stamp/stamp.jsp?tp=&arnumber=11218763}{Paper}&\href{https://github.com/MisEty/GPS-SLAM}{Code}\\

GS-LIVM\cite{GS-LIVM}& \multicolumn{1}{c|}{ICCV 2025} & \multicolumn{1}{c|}{F,R3} & \multicolumn{1}{c}{\checkmark} & \multicolumn{1}{c}{} & \multicolumn{1}{c}{\checkmark} & \multicolumn{1}{c|}{\checkmark} & \multicolumn{1}{c}{\checkmark}  & \multicolumn{1}{c}{}  & \multicolumn{1}{c}{}  & \multicolumn{1}{c|}{}  & \multicolumn{1}{c}{} & \multicolumn{1}{c|}{\checkmark} & \multicolumn{1}{c|}{} &\href{https://arxiv.org/pdf/2410.17084}{Paper}&\href{https://github.com/xieyuser/GS-LIVM}{Code}\\

\rowcolor[HTML]{ECF4FF}
S3PO-GS\cite{S3PO-GS}          & \multicolumn{1}{c|}{ICCV 2025}& \multicolumn{1}{c|}{K,W} & \multicolumn{1}{c}{\checkmark} & \multicolumn{1}{c}{} & \multicolumn{1}{c}{} & \multicolumn{1}{c|}{} & \multicolumn{1}{c}{\checkmark}  & \multicolumn{1}{c}{}  & \multicolumn{1}{c}{}  & \multicolumn{1}{c|}{}  & \multicolumn{1}{c}{} & \multicolumn{1}{c|}{\checkmark} & \multicolumn{1}{c|}{} &\href{https://arxiv.org/pdf/2507.03737}{Paper}&\href{https://github.com/3DAgentWorld/S3PO-GS}{Code}\\

SEGS-SLAM\cite{SEGS-SLAM}          & \multicolumn{1}{c|}{ICCV 2025} & \multicolumn{1}{c|}{R,T,E} & \multicolumn{1}{c}{\checkmark} & \multicolumn{1}{c}{\checkmark} & \multicolumn{1}{c}{} & \multicolumn{1}{c|}{} & \multicolumn{1}{c}{\checkmark}  & \multicolumn{1}{c}{}  & \multicolumn{1}{c}{\checkmark}  & \multicolumn{1}{c|}{}  & \multicolumn{1}{c}{} & \multicolumn{1}{c|}{\checkmark} & \multicolumn{1}{c|}{} &\href{https://arxiv.org/pdf/2501.05242}{Paper}&\href{https://github.com/leaner-forever/SEGS-SLAM}{Code}\\

\rowcolor[HTML]{ECF4FF}
Gaussian-LIC2\cite{Gaussian-LIC2}& \multicolumn{1}{c|}{arxiv 2025} & \multicolumn{1}{c|}{F,R3} & \multicolumn{1}{c}{\checkmark} & \multicolumn{1}{c}{} & \multicolumn{1}{c}{\checkmark} & \multicolumn{1}{c|}{\checkmark} & \multicolumn{1}{c}{\checkmark}  & \multicolumn{1}{c}{}  & \multicolumn{1}{c}{}  & \multicolumn{1}{c|}{}  & \multicolumn{1}{c}{} & \multicolumn{1}{c|}{\checkmark} & \multicolumn{1}{c|}{} &\href{https://arxiv.org/pdf/2507.04004}{Paper}&\href{https://xingxingzuo.github.io/gaussian_lic2/}{Code}\\

GauS-SLAM \cite{GauS-SLAM}   & \multicolumn{1}{c|}{arXiv 2025} & \multicolumn{1}{c|}{R,T,S} & \multicolumn{1}{c}{} & \multicolumn{1}{c}{\checkmark} & \multicolumn{1}{c}{} & \multicolumn{1}{c|}{} & \multicolumn{1}{c}{\checkmark}  & \multicolumn{1}{c}{\checkmark}  & \multicolumn{1}{c}{}  & \multicolumn{1}{c|}{}  & \multicolumn{1}{c}{} & \multicolumn{1}{c|}{\checkmark} & \multicolumn{1}{c|}{} &\href{https://arxiv.org/pdf/2505.01934}{Paper}&\href{https://github.com/gaus-slam/gaus-slam}{Code}\\

\rowcolor[HTML]{ECF4FF}
VTGaussian-SLAM \cite{VTGaussian-SLAM}   & \multicolumn{1}{c|}{ICML 2025} & \multicolumn{1}{c|}{R,T,S} & \multicolumn{1}{c}{} & \multicolumn{1}{c}{\checkmark} & \multicolumn{1}{c}{} & \multicolumn{1}{c|}{} & \multicolumn{1}{c}{\checkmark}  & \multicolumn{1}{c}{}  & \multicolumn{1}{c}{}  & \multicolumn{1}{c|}{}  & \multicolumn{1}{c}{} & \multicolumn{1}{c|}{\checkmark} & \multicolumn{1}{c|}{} &\href{https://arxiv.org/pdf/2506.02741}{Paper}&\href{https://github.com/MachinePerceptionLab/VTGaussian-SLAM}{Code}\\

FGS-SLAM  \cite{FGS-SLAM}   & \multicolumn{1}{c|}{IROS 2025} & \multicolumn{1}{c|}{R,T} & \multicolumn{1}{c}{} & \multicolumn{1}{c}{\checkmark} & \multicolumn{1}{c}{} & \multicolumn{1}{c|}{} & \multicolumn{1}{c}{\checkmark}  & \multicolumn{1}{c}{\checkmark}  & \multicolumn{1}{c}{\checkmark}  & \multicolumn{1}{c|}{}  & \multicolumn{1}{c}{\checkmark} & \multicolumn{1}{c|}{} & \multicolumn{1}{c|}{} &\href{https://arxiv.org/pdf/2503.01109}{Paper}&\href{https://github.com/3DV-Coder/FGS-SLAM}{Code}\\

\rowcolor[HTML]{ECF4FF}
GS-SDF \cite{GS-SDF}   & \multicolumn{1}{c|}{IROS 2025} & \multicolumn{1}{c|}{R,F} & \multicolumn{1}{c}{\checkmark} & \multicolumn{1}{c}{} & \multicolumn{1}{c}{} & \multicolumn{1}{c|}{\checkmark} & \multicolumn{1}{c}{\checkmark}  & \multicolumn{1}{c}{}  & \multicolumn{1}{c}{}  & \multicolumn{1}{c|}{}  & \multicolumn{1}{c}{} & \multicolumn{1}{c|}{\checkmark} & \multicolumn{1}{c|}{} &\href{https://arxiv.org/pdf/2503.10170?}{Paper}&\href{https://github.com/hku-mars/GS-SDF}{Code}\\

KBGS-SLAM\cite{KBGS-SLAM}& \multicolumn{1}{c|}{SIVP 2025} & \multicolumn{1}{c|}{R,T} & \multicolumn{1}{c}{} & \multicolumn{1}{c}{\checkmark} & \multicolumn{1}{c}{} & \multicolumn{1}{c|}{} & \multicolumn{1}{c}{\checkmark}  & \multicolumn{1}{c}{\checkmark}  & \multicolumn{1}{c}{}  & \multicolumn{1}{c|}{}  & \multicolumn{1}{c}{} & \multicolumn{1}{c|}{\checkmark} & \multicolumn{1}{c|}{} &\href{https://link.springer.com/content/pdf/10.1007/s11760-025-04303-4.pdf?utm_source=clarivate&getft_integrator=clarivate}{Paper}&\href{}{-}\\ 

\rowcolor[HTML]{ECF4FF}
SFGS-SLAM\cite{SFGS-SLAM}& \multicolumn{1}{c|}{SCI-BASEL 2025} & \multicolumn{1}{c|}{R,T} & \multicolumn{1}{c}{} & \multicolumn{1}{c}{\checkmark} & \multicolumn{1}{c}{} & \multicolumn{1}{c|}{} & \multicolumn{1}{c}{}  & \multicolumn{1}{c}{}  & \multicolumn{1}{c}{\checkmark}  & \multicolumn{1}{c|}{}  & \multicolumn{1}{c}{\checkmark} & \multicolumn{1}{c|}{} & \multicolumn{1}{c|}{} &\href{https://www.mdpi.com/2076-3417/15/20/10876}{Paper}&\href{}{-}\\ 

\hline
\end{NiceTabular}}

\begin{minipage}{\linewidth}
\vspace{0.7ex}
\footnotesize
\setlength{\parindent}{0pt}  
\textbf{Notes:} For Dataset column, \textbf{R}=Replica; \textbf{T}=TUM; \textbf{S}=ScanNet; \textbf{E}=EuRoC; \textbf{F}=FAST-LIVO; \textbf{R3}=R3LIVE; \textbf{K}=KITTI; \textbf{MR}=MultiReplica; \textbf{W}=Waymo; \textbf{B}=Bonn. For Optimization Objective column, \textbf{RQ}=Rendering Quality (Sec.~\ref{3.1}); \textbf{TA}=Tracking Accuracy (Sec.~\ref{3.2}); \textbf{RS}=Reconstruction Speed (Sec.~\ref{3.3}); \textbf{MC}=Memory Consumption (Sec.~\ref{3.4}). For Tracking Strategy column, \textbf{F2F}=Frame-to-Frame; \textbf{F2M}=Frame-to-Model.
\end{minipage}
\end{table*}

\par\vspace{1ex}

\begin{table*}[]
\centering
\caption{Summary of common SLAM and 3DGS datasets and their characteristics}
\label{tab:all_datasets}
\setlength{\tabcolsep}{1pt}
\begin{tabular*}{\textwidth}{@{\extracolsep{\fill}}lcccccccc@{\extracolsep{\fill}}}
\toprule
\textbf{Dataset} & \textbf{Year} & \textbf{Pub.} & \textbf{Sensors} & \textbf{Source} & \textbf{Scene} & \textbf{Size} & \textbf{Location} & \textbf{Link} \\
\midrule
\multicolumn{9}{c}{\textbf{SLAM Datasets}} \\
\midrule
TUM RGB-D\cite{TUM} & 2012 & IROS & C, D & Real & Indoor & 39 sequences @30Hz & Germany &\href{https://vision.in.tum.de/data/datasets/rgbd-dataset}{web} \\
KITTI\cite{KITTI} & 2012 & CVPR & C, L, I & Real & Outdoor & 22 sequences @10Hz & Germany & \href{http://www.cvlibs.net/datasets/kitti}{web}  \\
ICL-NUIM\cite{ICL} & 2014 & ICRA & C, D & Sim & Indoor & 4 sequences @30Hz & UK/Ireland & \href{https://www.doc.ic.ac.uk/~ahanda/VaFRIC/iclnuim.html}{web}  \\
EuRoC MAV\cite{EuRoC} & 2016 & IJRR & C, I & Real & Indoor & 11 sequences @20Hz & Switzerland & \href{https://service.tib.eu/ldmservice/dataset/euroc-mav-dataset}{web}  \\
Oxford RobotCar\cite{Oxford} & 2017 & IJRR & C, L, I & Real & Outdoor & 100+ sequences & UK & \href{https://robotcar-dataset.robots.ox.ac.uk}{web} \\
ETH3D SLAM\cite{ETH} & 2019 & CVPR & C, D, I & Real & Indoor/Outdoor & 91 sequences @27Hz & Switzerland & \href{https://www.eth3d.net/}{web}  \\
Replica\cite{Replica} & 2019 & arXiv & C, D & Sim & Indoor & 90k images & USA & \href{https://github.com/facebookresearch/Replica-Dataset}{web}  \\
Bonn RGB-D\cite{Bonn} & 2019 & IROS & C, D & Real & Indoor & 26 sequences & Germany & \href{https://www.ipb.uni-bonn.de/data/rgbd-dynamic-dataset}{web}  \\
TartanAir\cite{TartanAir} & 2020 & IROS & C, D, L, I & Sim & Indoor/Outdoor & 100+ sequences & USA & \href{http://theairlab.org/tartanair-dataset}{web}  \\
Waymo Open\cite{Waymo} & 2020 & CVPR & C, L, I & Real & Outdoor & 1150 sequences @10Hz & USA & \href{https://waymo.com/open}{web} \\
OpenMPD\cite{OpenMPD} & 2022 & TVT & C, L & Real & Outdoor & 180 sequences @20Hz & China & \href{http://openmpd.com/}{web}  \\
FAST-LIVO\cite{FAST-LIVO} & 2022 & IROS & C, L, I & Real & Outdoor & 20 sequences & China & \href{https://github.com/hku-mars/FAST-LIVO}{web}  \\
R3LIVE\cite{R3LIVE} & 2022 & ICRA & C, L, I & Real & Indoor/Outdoor & 13 sequences @15Hz & China & \href{https://github.com/ziv-lin/r3live_dataset}{web}  \\
ScanNet++\cite{Scannet++} & 2023 & ICCV & C, D, L & Real & Indoor & 280k images & Germany & \href{https://scannetpp.mpi-inf.mpg.de}{web}  \\
MultiReplica\cite{MultiReplica} & 2023 & NeurIPS & C, D & Sim & Indoor & 16,800 images & USA & \href{https://zju3dv.github.io/cp-slam/}{web}  \\
\midrule
\multicolumn{9}{c}{\textbf{3DGS / Neural Rendering Datasets}} \\
\midrule
DTU MVS\cite{MVS_data} & 2014 & CVPR & C, D & Real & Indoor & 80 scenes, 49/64 images/scene & Denmark & \href{https://roboimagedata.compute.dtu.dk/?page_id=36}{web}  \\
Tanks and Temples\cite{Tanks} & 2017 & 3DV & C & Real & Indoor/Outdoor & 14 scenes & USA & \href{https://www.tanksandtemples.org}{web}  \\
RealEstate10K\cite{RealEstate} & 2018 & SIGGRAPH & C & Real & Indoor & 80k scenes, $\approx 10$M frames & USA & \href{https://google.github.io/realestate10k}{web}  \\
LLFF (nerf\_llff\_data)\cite{LLFF} & 2019 & TOG & C & Real & Indoor/Outdoor & 8 scenes, 20--62 images/scene & USA & \href{https://nerfbaselines.github.io/llff}{web}  \\
NeRF Synthetic (Blender)\cite{NeRF} & 2020 & ECCV & C & Sim & Object & 8 scenes, 400 images/scene & USA & \href{https://www.kaggle.com/datasets/nguyenhung1903/nerf-synthetic-dataset}{web}  \\
BlendedMVS\cite{BlendedMVS} & 2020 & CVPR & C, D & Sim/Real & Indoor/Outdoor & 113 scenes, 17k images & China & \href{https://github.com/YoYo000/BlendedMVS}{web}  \\
Mip-NeRF 360\cite{Mip-NeRF} & 2022 & CVPR & C & Real & Indoor/Outdoor & 9 scenes& USA & \href{https://jonbarron.info/mipnerf360}{web}  \\
UrbanScene3D\cite{UrbanScene3D} & 2022 & ECCV & C, L & Real/Sim & Urban-scale & 16 scenes, 128k images & China & \href{https://vcc.tech/UrbanScene3D}{web}  \\
Tandt\_db (3DGS)\cite{3DGS} & 2023 & SIGGRAPH & C & Real & Indoor/Outdoor & 4 scenes, 200--300 images/scene & France & \href{https://repo-sam.inria.fr/fungraph/3d-gaussian-splatting}{web}  \\
MatrixCity\cite{Matrixcity} & 2023 & ICCV & C, D & Sim & City-scale & 2 scenes, 519k images & China & \href{https://city-super.github.io/matrixcity}{web}  \\
\bottomrule
\end{tabular*}
\vspace{1ex}
\par\raggedright \footnotesize \textbf{Notes:} C: Camera, D: Depth/RGB-D, L: LiDAR, I: IMU.
\end{table*}

\textit{1)Initialization:}
On the first frame, the camera pose is set to the identity and tracking is skipped. During Gaussian initialization, one Gaussian is created per image pixel: its color is set to the pixel’s RGB value, its depth to the pixel’s measured depth, and its opacity $\alpha=0.5$. The 2D projected radius is fixed to one pixel, yielding a 3D Gaussian radius:
\begin{equation}
r = \frac{D_{\text{GT}}}{f},
\end{equation}
where $D_{\text{GT}}$ is the ground-truth depth and $f$ is the focal length. 

This provides an explicit initial scene representation for subsequent processing.

\textit{2)Camera Tracking:}
For each frame, a constant-velocity model predicts an initial pose:
\begin{equation}
E_{t+1} = E_t + (E_t - E_{t-1}),
\end{equation}
where  $E_t$ is the camera pose at time  $t$.

The pose is then optimized by minimizing a combined photometric-depth loss:
\begin{equation}
L_t = \sum_p \left( S(p) > 0.99 \right) \Big( L_1(D(p)) + 0.5 L_1(C(p)) \Big),
\end{equation}
where $L_1(D(p))$ and $L_1(C(p))$ are the $L_1$ depth and color losses at pixel $p$, and $S(p)$ is a “visibility score” indicating map reliability at $p$. The sum is over pixels with $S(p) > 0.99$, ensuring optimization uses only well-converged regions.

After tracking, frames that meet threshold are added to the keyframe queue for mapping.

\textit{3)Gaussian Mapping:}
Each new keyframe contributes to the Gaussian map.After obtaining the camera pose and depth information for each keyframe, a densification mask $M(p)$ is used to determine which regions require new Gaussians to compensate for insufficient coverage or foreground changes:
\begin{equation}
M(p) = \big(S(p) < 0.5\big) + \big(D_{\text{GT}}(p) < D(p)\big)\big(L_1(D(p)) > \lambda \,\text{MDE}\big),
\end{equation}
where $D_{\text{GT}}(p)$ is the ground-truth depth at pixel $p$, $D(p)$ is the predicted depth, $\text{MDE}$ denotes the median depth error, and $\lambda$ is an empirically chosen coefficient. For pixels that satisfy the mask conditions, new Gaussians are added in the same manner as during initialization, ensuring mapping quality without increasing computational overhead.

All Gaussians then undergo local joint optimization: their positions, scales, orientations, colors, and opacities are refined to minimize a combined photometric-depth loss:
\begin{equation}
L_g = \lambda_c \left\| C(p) - C_{\text{GT}}(p) \right\| + \lambda_d \left\| D(p) - D_{\text{GT}}(p) \right\|,
\end{equation}
where $\lambda_c$ and $\lambda_d$ weight the color and depth errors.

\textit{4)Loop Closure Optimization:}
When a loop closure is detected, a global pose-graph optimization is performed. Using the constructed 3DGS map as the basis, the poses of loop frames and their co-visible keyframes are fixed or jointly optimized, and the parameters of Gaussians in the loop region are re-optimized. This aligns the 3DGS map with all observations in the loop area, improving global consistency.

Fig.~\ref{4} illustrates the overall 3DGS-SLAM pipeline. Through these stages, 3DGS-SLAM systems combine SLAM’s robust pose estimation with 3DGS’s high-fidelity mapping to achieve real-time high-quality reconstruction.

\section{Performance Optimization of 3DGS-SLAM}
While 3DGS-SLAM brings new capabilities, it also introduces optimization challenges. Here, we survey recent works that aim to enhance the performance of 3DGS-SLAM systems across four key dimensions: \textbf{rendering quality} (Sec \ref{3.1}), \textbf{tracking accuracy} (Sec \ref{3.2}), \textbf{reconstruction speed} (Sec \ref{3.3}), and \textbf{memory consumption} (Sec \ref{3.4}). Table \ref{tab_1} presents a summary of representative optimization approaches in 3DGS-SLAM. Table \ref{tab:all_datasets} summarizes the datasets in this field.

\subsection{Rendering Quality}
\label{3.1}

\begin{figure*}[!t]
  \centering
  \includegraphics[width=\textwidth]{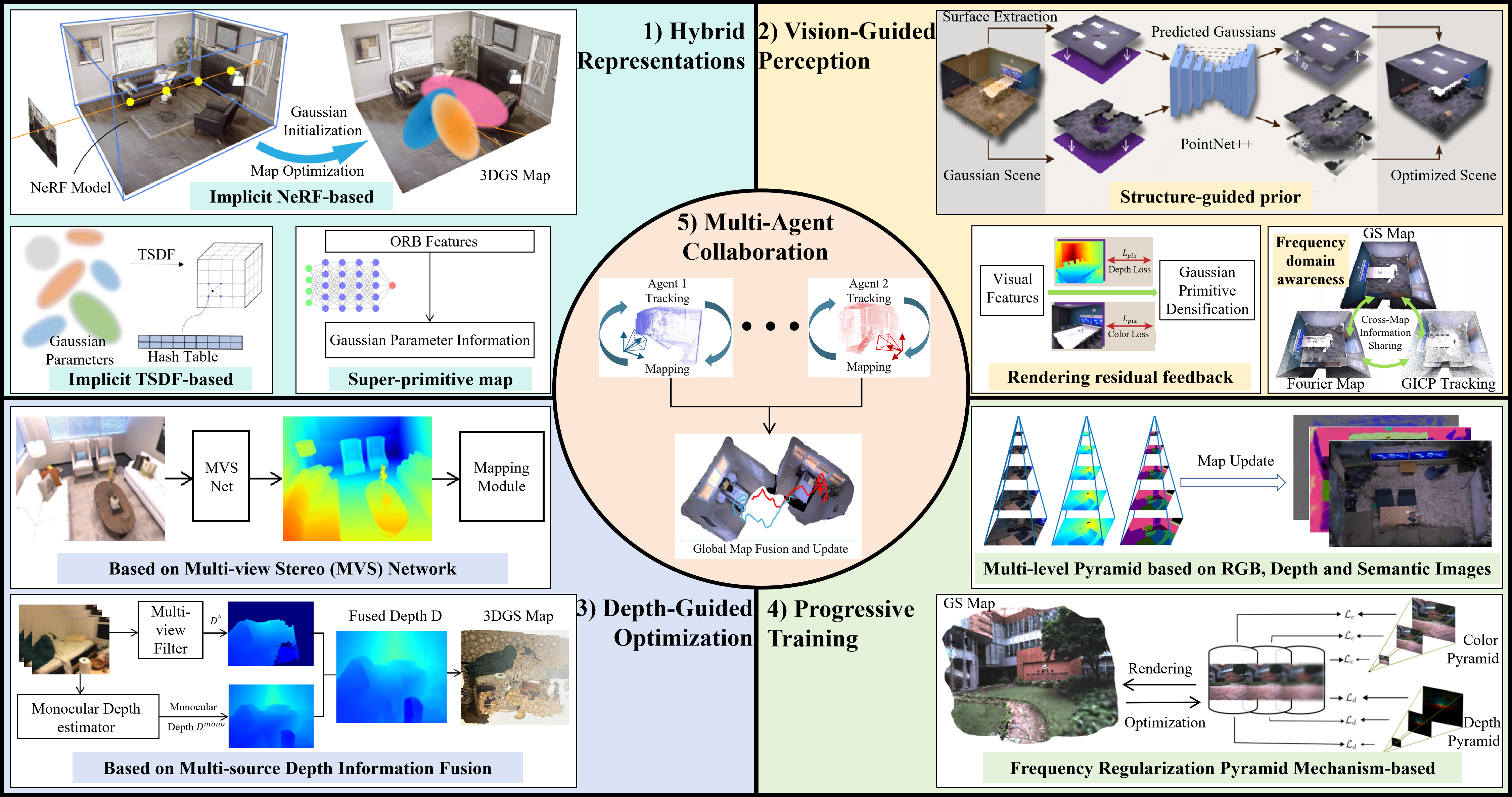} 
  \caption{Summary of rendering quality optimization methods. We categorize representative approaches into five strategies: 1) Hybrid Representations: combining explicit Gaussians with implicit priors for robust initialization; 2) Vision-Guided Perception: exploiting visual residuals and structural cues for primitive densification; 3) Depth-Guided Optimization: enhancing geometric accuracy via MVS or multi-source depth fusion; 4) Progressive Training: utilizing pyramid-based mechanisms for coarse-to-fine refinement; and 5) Multi-Agent Collaboration: facilitating global map fusion across distributed agents. }
  \label{5}
\end{figure*}

In 3DGS-SLAM, rendering quality is a primary metric for reconstruction, since high-quality rendering preserves fine scene details crucial for AR/VR. Early work\cite{MonoGS} demonstrated that 3DGS can achieve high-fidelity reconstruction in SLAM, but SLAM conditions (sparse views, scale ambiguity, missing depth) tend to degrade quality. To address this, many methods have been proposed. As shown in Fig.~\ref{5}, we categorize them into five groups, for each group we dissect the flagship techniques, quantify their specific contributions to rendering fidelity, and aggregate their perceptual–-metric performance on public benchmarks. Table \ref{tab1} compares representative methods, highlighting their strengths and limitations. Each category has advanced rendering quality in 3DGS-SLAM under different conditions, addressing issues like sparse inputs, unobserved areas, and texture detail.

\renewcommand{\arraystretch}{1.2}
\begin{table*}[h]
\caption{Comparison of Rendering Quality Optimization Methods in 3DGS-SLAM}
\label{tab1}
\centering
\newcolumntype{L}[1]{>{\raggedright\arraybackslash}p{#1}} 

\begin{tabular*}{\textwidth}{@{\extracolsep{\fill}}
  L{2.0cm}  
  L{3.3cm}  
  L{4.2cm}  
  L{3.5cm}  
  L{3cm}  
@{\extracolsep{\fill}}}
\toprule
\textbf{Category} & \textbf{Representative Methods} & \textbf{Advantages} & \textbf{Limitations} & \textbf{Typical Scenarios} \\
\midrule

\multirow{3}{=}{Hybrid Explicit-Implicit Representations} 
& Photo\cite{Photo}, NGM\cite{NGM-SLAM}, DenseSplat\cite{DenseSplat}, Constrained\cite{Constrained} 
& Combines explicit geometry with implicit priors to fill sparse areas and enhance texture fidelity. 
& Higher computational cost; relies on alignment between representations. 
& Scenes with sparse observations or surface completion needs. \\ 
\midrule

\multirow{4}{=}{Vision-Guided Perception} 
& HF-SLAM\cite{HF-SLAM}, Gaussian-SLAM\cite{Gaussian-SLAM}, MG\cite{MG-SLAM}, 2DGS-SLAM\cite{2DGS-SLAM}, SEGS\cite{SEGS-SLAM}, FGS\cite{FGS-SLAM} 
& Uses visual cues (residuals, structural/frequency priors) for adaptive densification and improved structural coherence. 
& Structural priors limit generalization; residual methods struggle with occlusions. 
& Indoor architectural environments or scenes with variable texture frequency. \\ 
\midrule

\multirow{4}{=}{Depth-Guided Optimization} 
& Splat\cite{Splat-SLAM}, DROID\cite{DROID-Splat}, DP\cite{DP-SLAM}, MGS\cite{MGS-SLAM}, MVS-GS\cite{MVS-GS}, SplatMAP\cite{SplatMAP}, GauS\cite{GauS-SLAM}, VTGaussian\cite{VTGaussian-SLAM} 
& Regularizes geometry via multi-source depth priors and uncertainty weighting, minimizing artifacts in textureless regions. 
& Heavily dependent on the accuracy of external depth priors; sensitive to sensor noise. 
& Textureless regions, dense capture, or monocular setups needing geometric constraints. \\ 
\midrule

\multirow{3}{=}{Progressive Training} 
& Scaffold\cite{Scaffold-SLAM}, MotionGS\cite{MotionGS}, LVI-GS\cite{LVI-GS}, RGBDS\cite{RGBDS-SLAM}, FIGS\cite{FIGS-SLAM} 
& Coarse-to-fine optimization prevents early overfitting and ensures global consistency. 
& Complex multi-stage pipeline; high-frequency refinement requires high-quality data. 
& Large-scale scenes requiring stable convergence. \\ 
\midrule

\multirow{3}{=}{Multi-Agent Collaboration} 
& \multirow{3}{=}{MAGiC-SLAM\cite{MAGiC-SLAM}, GRAND-SLAM\cite{GRAND-SLAM} }
& Enables scalable global mapping via submap fusion; optimizes bandwidth via visibility masking. 
& High coordination overhead; relies on robust loop closure for map merging. 
& Large-scale distributed exploration or multi-robot systems. \\ 

\bottomrule
\end{tabular*}
\end{table*}

\textit{1) Hybrid Explicit-Implicit Representations:} To overcome the limitations of discrete primitives in capturing continuous surfaces and fine details, these methods integrate explicit Gaussians with implicit neural representations. Approaches such as Photo-SLAM\cite{Photo}, NGM-SLAM\cite{NGM-SLAM}, and DenseSplat\cite{DenseSplat} utilize neural fields or NeRF submaps to supervise Gaussian attributes, effectively filling in sparsely scanned areas and enhancing texture fidelity via volumetric rendering. Beyond neural supervision, Li et al.\cite{Constrained} incorporate geometric priors by employing a multi-resolution hash grid to predict TSDF values, jointly optimizing explicit parameters and implicit rendering losses to ensure geometric consistency.

\textit{2) Vision-Guided Perception:} Visual cues—ranging from rendering residuals to structural and frequency priors—are exploited to guide the adaptive densification and placement of Gaussians. Residual-based strategies, including HF-SLAM\cite{HF-SLAM} and Gaussian-SLAM\cite{Gaussian-SLAM}, drive optimization by monitoring color and depth errors to identifying under-reconstructed regions. To improve structural coherence, methods like MG-SLAM\cite{MG-SLAM}, 2DGS-SLAM\cite{2DGS-SLAM}, and SEGS-SLAM\cite{SEGS-SLAM} leverage geometric constraints, such as Manhattan-world assumptions, 2D planar compression, or point cloud anchors from ORB-SLAM3\cite{ORB-SLAM3}. Alternatively, FGS-SLAM\cite{FGS-SLAM} adopts a frequency-domain perspective, applying high-pass filtering to densely initialize Gaussians in texture-rich areas while maintaining sparsity in low-frequency regions.

\textit{3) Depth-Guided Optimization:} Accurate depth supervision is critical for regularizing geometry and minimizing artifacts, particularly in textureless or noisy regions. Several frameworks enhance 3DGS by fusing multi-source depth priors: Splat-SLAM\cite{Splat-SLAM} and DROID-Splat\cite{DROID-Splat} combine monocular predictions with multi-view or pseudo RGB-D cues. Others, such as MGS-SLAM\cite{MGS-SLAM} and MVS-GS\cite{MVS-GS}, rely on Multi-View Stereo (MVS) networks to generate dense depth maps for initialization. To address noise and distortion in these priors, recent works introduce uncertainty weighting (SplatMAP\cite{SplatMAP}) or geometric constraints like 2D surfels and visibility masks (GauS-SLAM\cite{GauS-SLAM}, VTGaussian-SLAM\cite{VTGaussian-SLAM}), ensuring robust updates under viewpoint variations.
 
\textit{4) Progressive Training:} To ensure global consistency while recovering fine details, progressive training strategies adopt hierarchical or multi-scale optimization frameworks. Methods such as Photo-SLAM\cite{Photo}, NGM-SLAM\cite{NGM-SLAM}, and LVI-GS\cite{LVI-GS} utilize image pyramids or adaptive voxel merging to refine the reconstruction from coarse global structures to fine local textures. Similarly, frequency-domain approaches like Scaffold-SLAM\cite{Scaffold-SLAM} and FIGS-SLAM\cite{FIGS-SLAM} prioritize stable low-frequency information before gradually resolving high-frequency details, often aided by pruning mechanisms to prevent premature convergence to local minima. Extending this further, RGBDS-SLAM\cite{RGBDS-SLAM} incorporates semantic pyramids into the joint optimization to preserve semantic consistency across scales.

\textit{5)Multi-Agent Collaboration:} Collaborative systems improve reconstruction scalability and fidelity by fusing submaps from multiple agents into a unified global representation. These methods focus on robust data integration and efficient communication. MAGiC-SLAM\cite{MAGiC-SLAM} partitions the scene into submaps and fuses submaps generated by different agents into a unified global map via loop closure detection, using rendering residual masks to filter unreliable regions and optimizing bandwidth by synchronizing only non-visible Gaussians. GRAND-SLAM\cite{GRAND-SLAM} adopts a local submap optimization strategy in which each agent independently refines Gaussian parameters using a mixed L1 loss weighted by color and depth cues. An outlier mask is further applied to suppress unstable pixel regions, enabling cross-scene, high-fidelity reconstructions across diverse datasets.

\begin{table}[]
\caption{Rendering Quality Evaluation on Replica Dataset}
\label{tab2}
\centering
\scriptsize
\setlength{\tabcolsep}{2pt}
\begin{NiceTabular}{l|ccc|ccc}
\noalign{\hrule height 1pt}
                          & \multicolumn{3}{c|}{Replica} & \multicolumn{3}{c}{TUM} \\ \cline{2-7} 
\multirow{-2}{*}{Methods} & \textbf{PSNR↑} & \textbf{SSIM↑} & \textbf{LPIPS↓}   & \textbf{PSNR↑} & \textbf{SSIM↑} & \textbf{LPIPS↓}  \\ \hline
    SplaTAM\cite{SplaTAM} & 34.11  & 0.970 & 0.100 & 22.80  &  0.893 &0.178\\
    Photo-SLAM (RGB)\cite{Photo}  & 33.30  & 0.926 & 0.078 & 20.55  &  0.720 &0.211\\
    Photo-SLAM (RGBD)\cite{Photo} & 34.96  & 0.942 & 0.059 &  21.90 & 0.763  &0.187\\
    NGM-SLAM (RGB)\cite{NGM-SLAM}  & 35.02  & 0.960 & 0.130&  -      &    -    &- \\
    NGM-SLAM (RGBD)\cite{NGM-SLAM} & 37.43  & 0.980 & 0.080 &    -    &    -    &-\\
    MonoGS\cite{MonoGS} & 38.94  & 0.968 & 0.070&   24.37     &  0.804      & 0.225\\
    Gaussian-SLAM\cite{Gaussian-SLAM} & \cellcolor{orange!50}42.08  & 0.996 & \cellcolor{orange!50}0.018&  25.05      &  \cellcolor{yellow!30}0.929      & 0.168 \\
    MVS-GS\cite{MVS-GS}           & 35.58  & 0.960 & 0.080 &   22.52     &   0.810     &0.210\\
    MG-SLAM\cite{MG-SLAM}         & 33.59  & 0.930 & 0.220&    -    &    -    &- \\
    MGS-SLAM\cite{MGS-SLAM}       & 32.41  & 0.918 & 0.088 & -       & -       &-\\
    SplatMAP\cite{SplatMAP}       & 33.93  & 0.974 & 0.064 &  23.12      & 0.879       &0.196\\
    GS-Loop\cite{GS-Loop}   &  37.96  & 0.987 & 0.051&     -   & -       &-\\
    DP-SLAM\cite{DP-SLAM}   & -  & - & -&   21.53     &  0.861     &0.205\\
    GS3LAM\cite{GS3LAM} & 36.26  & 0.989 & 0.052&    -    &     -   &- \\
    Splat-SLAM\cite{Splat-SLAM}   & 36.45  & 0.950 & 0.060 &  25.85      &    0.810    &0.190\\
    HF-SLAM\cite{HF-SLAM}   & 36.19  & 0.980 & 0.050 &   22.30     &  0.890      &0.160\\
    DenseSplat\cite{DenseSplat}   & 38.73  & 0.969 & 0.056 &   -     &    -    &-\\
    RGBDS-SLAM\cite{RGBDS-SLAM}   & 38.85  & 0.967 & 0.035&  -      &   -     & -\\
    Scaffold-SLAM(RGB)\cite{Scaffold-SLAM} & 37.71  & 0.963 & 0.041&  24.52     &  0.823      & 0.153\\
    Scaffold-SLAM(RGBD)\cite{Scaffold-SLAM} & 39.14  & 0.974 & 0.023&  25.95      &  0.853      & 0.160\\
    MotionGS\cite{MotionGS} & 39.60  & 0.976 & 0.043&    -    &     -   &- \\
    FlashSLAM\cite{FlashSLAM} & 39.21  & 0.976 & 0.042 &  22.85  & -    & -\\
    DROID-Splat(RGB)\cite{DROID-Splat} & 39.47  & 1.000 & 0.030&  \cellcolor{orange!50}26.84      & 0.990       & 0.130\\
    DROID-Splat(RGBD)\cite{DROID-Splat} & 39.66  & \cellcolor{red!50}1.000 & 0.030&   26.81     &   \cellcolor{orange!50}0.990    & \cellcolor{yellow!30}0.120 \\
    GSFF-SLAM\cite{GSFF-SLAM} & 38.67  & 0.974 & 0.035&    20.57    &0.736 &0.311 \\
    Constrained-SLAM\cite{Constrained} & 35.55  & 0.980 & 0.080&    -    &    -    &- \\
    SemGauss-SLAM\cite{SemGauss-SLAM} & 35.03  & 0.982 & 0.062&    -    &    -    &- \\
    2DGS-SLAM\cite{2DGS-SLAM}     & 38.50  & 0.972 & 0.045 &   -     &     -   &-\\
    GS4\cite{GS4}   & -  & - & -&   22.70     &  0.903     &0.191\\
    FGO-SLAM(RGB)\cite{FGO-SLAM} & 34.13  & 0.956 & 0.094 &     -   &    -    &-\\
    FGO-SLAM(RGBD)\cite{FGO-SLAM} & 38.35  & 0.973 & 0.084 &     -   &    -    &-\\
    FIGS-SLAM\cite{FIGS-SLAM} & 39.36  & 0.975 & 0.046 &  24.52    & 0.858       &0.198\\
    FGS-SLAM\cite{FGS-SLAM} & 38.75  & 0.974 & 0.041 &   -     &   -     &-\\
    SEGS-SLAM(RGB)\cite{SEGS-SLAM} & 37.96  & 0.964 & 0.037 &  25.17    & 0.825  & 0.122\\
    SEGS-SLAM(RGBD)\cite{SEGS-SLAM} & 39.42  & 0.975 & \cellcolor{yellow!30}0.021 &   \cellcolor{yellow!30}26.03     & 0.843   &\cellcolor{orange!50}0.107\\
    KBGS-SLAM\cite{KBGS-SLAM}& 39.34  & 0.975 & 0.043 &    -    & -       &-\\
    GauS-SLAM\cite{GauS-SLAM} & \cellcolor{yellow!30}40.25  & \cellcolor{yellow!30}0.991 & 0.027 &  25.45      & 0.922       &0.170\\
    VTGaussian-SLAM\cite{VTGaussian-SLAM} & \cellcolor{red!50}43.34  &\cellcolor{orange!50}0.996 & \cellcolor{red!50}0.012&  \cellcolor{red!50}30.20      &   \cellcolor{red!50}0.972     &\cellcolor{red!50}0.062 \\

\noalign{\hrule height 1pt}
\end{NiceTabular}
\end{table}

In conclusion, this section reviews five core optimization strategies designed to address rendering degradation in 3DGS-SLAM systems: \textbf{hybrid explicit-implicit representations} integrate explicit Gaussian voxels with implicit neural fields to enhance geometric consistency while preserving high-frequency details; \textbf{vision-guided perception} leverages frequency-domain analysis, structural priors, and rendering residual feedback to achieve adaptive Gaussian initialization and optimization; \textbf{depth-guided optimization} improve geometric accuracy and depth fidelity through multi-source depth fusion, edge-weighted depth constraints, and surface-aware rendering; \textbf{progressive training} paradigms employ multi-scale hierarchical strategies that recover scene spectrum features through coarse-to-fine optimization across different pyramid levels; \textbf{multi-agent collaborative} integrates distributed submap optimization, local geometric constraints, and dynamic voxel synchronization.

Table \ref{tab2} summarizes the rendering quality of representative methods on the Replica dataset. Throughout this paper, in all tables involving quantitative evaluations, we highlight the \colorbox{red!50}{best}, \colorbox{orange!50}{second-best}, and \colorbox{yellow!30}{third-best} results in red, orange, and yellow, respectively. Collectively, these advancements optimize 3DGS-SLAM from the perspectives of representation, geometry, perception, training, and system architecture, leading to significant improvements in rendering metrics such as PSNR and SSIM on public benchmarks.

\subsection{Tracking Accuracy}
\label{3.2}
Tracking accuracy is crucial for 3DGS-SLAM’s stability and map reliability. High-precision pose estimation ensures accurate map construction for AR and robotic tasks. Although 3DGS’s efficient rendering aids real-time mapping, SLAM challenges like fast motion, low-texture areas, and dynamic interference still cause pose drift. Some systems suffer from error accumulation due to lacking effective local/global optimization that exploits spatio-temporal data. To improve tracking, researchers have explored three categories of methods: local optimization, global pose-graph optimization, and global bundle-adjustment (BA) optimization. This section provides an overview of these three categories of methods, outlining their core concepts, key techniques, and interrelations.

\textit{1) Local Optimization:} These methods aim to minimize short-term drift by refining poses within limited spatial or temporal windows. A common strategy involves window-based joint optimization: OGS-SLAM\cite{OGS-SLAM} and FGS-SLAM\cite{FGS-SLAM} construct local co-visibility maps or dynamic keyframe windows to jointly optimize camera poses and map consistency. To further constrain the optimization, other approaches integrate geometric and depth priors. For instance, MGS-SLAM\cite{MGS-SLAM} and SplatMAP\cite{SplatMAP} introduce scale synchronization and depth-smoothness regularizers, respectively, while MG-SLAM\cite{MG-SLAM} explicitly incorporates line segments and plane priors to reduce geometric errors. Additionally, adaptive Gaussian management plays a crucial role in stabilizing tracking: RTG-SLAM\cite{RTG-SLAM} focuses computational effort by optimizing only “unstable” Gaussians, DenseSplat\cite{DenseSplat} applies adaptive density control, and GauS-SLAM\cite{GauS-SLAM} employs a periodic frame-to-model registration reset to prevent drift accumulation.

\begin{figure}[]
\centering
\begin{tikzpicture}
    \definecolor{color1}{rgb}{0.12, 0.46, 0.70}
    \definecolor{color2}{rgb}{0.89, 0.10, 0.11}
    \definecolor{color3}{rgb}{0.60, 0.31, 0.64}
    \definecolor{color4}{rgb}{0.30, 0.59, 0.38}
    \definecolor{color5}{rgb}{0.99, 0.68, 0.38}
    \definecolor{color6}{rgb}{0.65, 0.81, 0.89}
    \definecolor{color7}{rgb}{0.70, 0.87, 0.54}
    \definecolor{color8}{rgb}{0.29, 0.88, 0.60}
    \definecolor{color9}{rgb}{0.80, 0.50, 0.00}
    \definecolor{color10}{rgb}{0.60, 0.20, 0.80}
    \definecolor{color11}{rgb}{0.00, 0.60, 0.60}
    \definecolor{color12}{rgb}{0.90, 0.70, 0.00}
    \definecolor{color13}{rgb}{0.40, 0.40, 0.40}
    \definecolor{color14}{rgb}{0.7, 0.7, 0.7}
    \definecolor{color15}{rgb}{0.9, 0.5, 0.1}
    \definecolor{color16}{rgb}{0.2, 0.1, 0.5}
    
    \definecolor{color17}{rgb}{0.9, 0.1, 0.5}
    \definecolor{color18}{rgb}{0.1, 0.3, 0.9}
    \definecolor{color19}{rgb}{0.5, 0.1, 0.9}
    \definecolor{color20}{rgb}{0.4, 0.8, 0.5}
    \definecolor{color21}{rgb}{0.55, 0.1, 0.6}
    \definecolor{color22}{rgb}{0.15, 0.25, 0.1}

     \definecolor{color23}{rgb}{0.75, 0.25, 0.1}

    \definecolor{color24}{rgb}{0.15, 0.05, 0.95}
        \definecolor{color25}{rgb}{0.75, 0.05, 0.45}

            \definecolor{color26}{rgb}{0.55, 0.45, 0.25}
    \pgfmathsetmacro{\newmarksize}{1.5}

    \begin{axis}[
        name=main plot,
        xlabel={Replica ATE RMSE [cm]$\downarrow$},
        ylabel={TUM ATE RMSE [cm]$\downarrow$},
        xlabel style={font=\footnotesize},
        ylabel style={font=\footnotesize},
        xmin=0, xmax=9,
        ymin=0, ymax=9,
        ytick={0,1,2,3,4,5,6,7,8,9},
        yticklabels={-,1,2,3,4,5,6,7,8,9},
        xtick={0,1,2,3,4,5,6,7,8,9},
        grid=major,
        grid style={dashed, gray!30},
        width=9.4cm,
        height=8cm
    ]

    \addplot[mark=*, mark size=\newmarksize pt, color=color1] coordinates {(0.32, 2.93)};
    \addplot[mark=*, mark size=\newmarksize pt, color=color2] coordinates {(0.33, 1.58)};
    \addplot[mark=triangle*, mark size=\newmarksize pt, color=color3] coordinates {(0.32, 5.37)};
    \node[above right, text=color3, font=\scriptsize] at (axis cs:0.32,5.37) {Mon-SLAM};

    \addplot[mark=triangle*, mark size=\newmarksize pt, color=color4] coordinates {(0.23, 2.23)};

    \addplot[mark=square*, mark size=\newmarksize pt, color=color5] coordinates {(0.26, 0.00)};

    \addplot[mark=square*, mark size=\newmarksize pt, color=color6] coordinates {(8.51, 0.00)};
    \node[above left, text=color6, font=\scriptsize] at (axis cs:8.51,0.00) {NGM-SLAM (RGB)};

    \addplot[mark=square*, mark size=\newmarksize pt, color=color7] coordinates {(0.27, 1.8)};

    \addplot[mark=*, mark size=\newmarksize pt, color=color8] coordinates {(0.36, 3.81)};

    \addplot[mark=*, mark size=\newmarksize pt, color=color9] coordinates {(0.52, 8.92)};
    \node[below right, text=color9, font=\scriptsize] at (axis cs:0.52,8.92) {Point-SLAM};

    \addplot[mark=square*, mark size=\newmarksize pt, color=color11] coordinates {(0.29, 1.9)};

    \addplot[mark=triangle*, mark size=\newmarksize pt, color=color12] coordinates {(0.26, 3.33)};

    \addplot[mark=square*, mark size=\newmarksize pt, color=color13] coordinates {(0.29, 1.33)};

    \addplot[mark=*, mark size=\newmarksize pt, color=color14] coordinates {(0.15, 2.00)};

    \addplot[mark=*, mark size=\newmarksize pt, color=color15] coordinates {(0.06, 1.54)};

    \addplot[mark=*, mark size=\newmarksize pt, color=color16] coordinates {(0.45, 0.00)};
    \addplot[mark=*, mark size=\newmarksize pt, color=color17] coordinates {(0.38, 0.91)};
    \addplot[mark=*, mark size=\newmarksize pt, color=color18] coordinates {(0.00, 1.4)};
      \addplot[mark=*, mark size=\newmarksize pt, color=color19] coordinates {(0.28, 0)};
    \addplot[mark=*, mark size=\newmarksize pt, color=color20] coordinates {(0, 5.05)};
     \node[below right, text=color20, font=\scriptsize] at (axis cs:0, 5.05) {OGS-SLAM};
               
    \addplot[mark=*, mark size=\newmarksize pt, color=color21] coordinates {(0.21, 1.93)};

     \addplot[mark=*, mark size=\newmarksize pt, color=color22] coordinates {(0.27, 2.55)};

     \addplot[mark=*, mark size=\newmarksize pt, color=color23] coordinates {(0.28, 1.49)};
    \addplot[mark=*, mark size=\newmarksize pt, color=color24] coordinates {(0.29,3.85)};
     \addplot[mark=*, mark size=\newmarksize pt, color=color25] coordinates {(0,2.72)};
          \addplot[mark=*, mark size=\newmarksize pt, color=color26] coordinates {(0,0.98)};
    \draw[solid, thick, black] (axis cs:0, 0) rectangle (axis cs:0.5, 4);
    \coordinate (main_arrow_start) at (axis cs:0.5, 2);

    \end{axis}

    \begin{axis}[
        name=inset plot,
        at={(2cm, 0.8cm)},
        anchor=south west,
        width=7cm,
        height=6.4cm,
        xmin=0, xmax=0.5,
        ymin=0, ymax=4,
        ytick={0,0.5,1,1.5,2,2.5,3,3.5,4},
        yticklabels={-,0.5,1,1.5,2,2.5,3,3.5,4},
        xtick={0,0.1,0.2,0.3,0.4,0.5},
        grid=major,
        grid style={dashed, gray!30},
        axis background/.style={fill=gray!10},
        title={Zoomed-in View},
        title style={font=\tiny},
        xlabel={},
        ylabel={},
        xlabel style={font=\tiny},
        ylabel style={font=\tiny},
        xticklabel style={font=\tiny},
        yticklabel style={font=\tiny}
    ]

    \addplot[mark=*, mark size=\newmarksize pt, color=color1] coordinates {(0.32, 2.93)};
    \node[right, text=color1, font=\tiny] at (axis cs:0.32,2.93) {MGS-SLAM};

    \addplot[mark=*, mark size=\newmarksize pt, color=color2] coordinates {(0.33, 1.58)};
    \node[right, text=color2, font=\tiny] at (axis cs:0.33,1.58) {DenseSplat};

    \addplot[mark=triangle*, mark size=\newmarksize pt, color=color4] coordinates {(0.23, 2.23)};
    \node[left, text=color4, font=\tiny] at (axis cs:0.23,2.23) {GLC-SLAM};

    \addplot[mark=square*, mark size=\newmarksize pt, color=color5] coordinates {(0.26, 0.00)};
    \node[above left, text=color5, font=\tiny] at (axis cs:0.26,0.00) {HI-SLAM2};

    \addplot[mark=square*, mark size=\newmarksize pt, color=color7] coordinates {(0.27, 1.8)};
    \node[right, text=color7, font=\tiny] at (axis cs:0.15,1.7) {DROID-Splat (RGB)};

    \addplot[mark=*, mark size=\newmarksize pt, color=color8] coordinates {(0.36, 3.81)};
    \node[right, text=color8, font=\tiny] at (axis cs:0.36,3.81) {SplaTAM};

    \addplot[mark=square*, mark size=\newmarksize pt, color=color11] coordinates {(0.29, 1.9)};
    \node[right, text=color11, font=\tiny] at (axis cs:0.29, 1.9) {DROID-Splat (RGB-D)};

    \addplot[mark=triangle*, mark size=\newmarksize pt, color=color12] coordinates {(0.26, 3.33)};
    \node[above right, text=color12, font=\tiny] at (axis cs:0.26,3.33) {LoopSplat};

    \addplot[mark=square*, mark size=\newmarksize pt, color=color13] coordinates {(0.29, 1.33)};
    \node[right, text=color13, font=\tiny] at (axis cs:0.29,1.33) {Constrained-SLAM};

    \addplot[mark=*, mark size=\newmarksize pt, color=color14] coordinates {(0.15, 2.00)};
    \node[left, text=color14, font=\tiny] at (axis cs:0.15,2.00) {FGS-SLAM};

    \addplot[mark=*, mark size=\newmarksize pt, color=color15] coordinates {(0.06, 1.54)};
    \node[right, text=color15, font=\tiny] at (axis cs:0.06,1.54) {GauS-SLAM};

    \addplot[mark=*, mark size=\newmarksize pt, color=color16] coordinates {(0.45, 0.00)};
    \node[above, text=color16, font=\tiny] at (axis cs:0.45,0.00) {MG-SLAM};

    \addplot[mark=*, mark size=\newmarksize pt, color=color17] coordinates {(0.38, 0.91)};
    \node[below, text=color17, font=\tiny] at (axis cs:0.38,0.91) {GSORB-SLAM};
    
    \addplot[mark=*, mark size=\newmarksize pt, color=color18] coordinates {(0.00, 1.40)};
    \node[right, text=color18, font=\tiny] at (axis cs:0.00, 1.40) {FT-SLAM};

    \addplot[mark=*, mark size=\newmarksize pt, color=color19] coordinates {(0.28, 0)};
    \node[above right, text=color19, font=\tiny] at (axis cs:0.28, 0) {DSOSplat};

    \addplot[mark=*, mark size=\newmarksize pt, color=color20] coordinates {(0, 5.05)};
    \node[above right, text=color20, font=\tiny] at (axis cs:0, 5.05) {OGS-SLAM};

        \addplot[mark=*, mark size=\newmarksize pt, color=color21] coordinates {(0.21, 1.93)};
    \node[above, text=color21, font=\tiny] at (axis cs:0.21, 1.93) {S3LAM};

            \addplot[mark=*, mark size=\newmarksize pt, color=color22] coordinates {(0.27, 2.55)};
    \node[right, text=color22, font=\tiny] at (axis cs:0.27, 2.55) {KBGS-SLAM};

                \addplot[mark=*, mark size=\newmarksize pt, color=color23] coordinates {(0.28, 1.49)};
    \node[left, text=color23, font=\tiny] at (axis cs:0.28, 1.49) {GS-Loop};

                \addplot[mark=*, mark size=\newmarksize pt, color=color24] coordinates {(0.29, 3.85)};
    \node[left, text=color24, font=\tiny] at (axis cs:0.29, 3.85) {Loopy-SLAM};

                    \addplot[mark=*, mark size=\newmarksize pt, color=color25] coordinates {(0, 2.72)};
    \node[right, text=color25, font=\tiny] at (axis cs:0, 2.72) {GI-SLAM};

     \addplot[mark=*, mark size=\newmarksize pt, color=color26] coordinates {(0, 0.98)};
    \node[right, text=color26, font=\tiny] at (axis cs:0, 0.98) {FGO-SLAM};

    \coordinate (inset_arrow_end) at (axis cs:0, 2);
    \end{axis}

    \draw[black, thick, -stealth] (main_arrow_start) -- (inset_arrow_end);

\end{tikzpicture}

\caption{Comparison of tracking accuracy on the Replica and TUM datasets. Some methods did not produce results on the corresponding datasets, and thus their values are missing in the figure.}
    \label{6}
\end{figure}
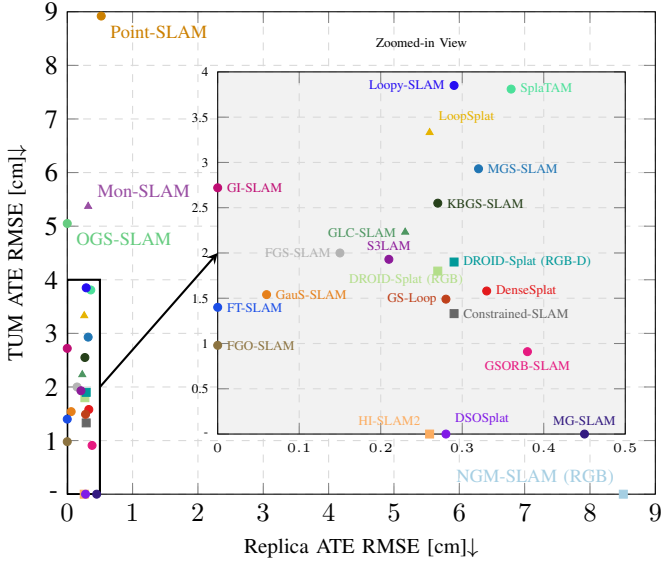

\textit{2)Global Pose-Graph Optimization:}
To correct accumulated long-term drift, these methods construct a global graph to enforce consistency across keyframes or submaps, primarily triggered by loop closure detection. Traditional feature-based approaches, such as GS-Loop\cite{GS-Loop} and GLC-SLAM\cite{GLC-SLAM}, rely on ORB features or visual-geometric overlap to identify loops and optimize the pose graph using solvers like g2o or Levenberg–Marquardt. In contrast, recent methods leverage deep learning-based descriptors for more robust place recognition. LoopSplat\cite{LoopSplat}, MAGiC-SLAM\cite{MAGiC-SLAM}, and Mon-SLAM\cite{Mon-SLAM} utilize advanced embeddings—such as NetVLAD, DinoV2, and CLIP—to detect loops even with significant viewpoint changes, subsequently optimizing weighted pose graphs or neural networks to rigidly align submaps and keyframes globally. These graph optimizations enforce global consistency across all keyframes.

\textit{3)Global BA Optimization:} While pose-graph optimization constrains poses, it may not correct all geometric drift. Therefore, some works introduce Global BA to jointly refine both camera poses and the 3DGS map geometry (Gaussian parameters) to maximize photometric and geometric consistency. Several frameworks adopt factor graph formulations with depth priors: Splat-SLAM\cite{Splat-SLAM} and DROID-Splat\cite{DROID-Splat} integrate monocular depth predictions or disparity terms into the factor graph, enabling the simultaneous optimization of poses, intrinsics, and scale. Multi-stage and multi-modal strategies are also employed to handle complex scenes; HI-SLAM2\cite{HI-SLAM2} combines online Sim(3) optimization with offline global refinement, while NGM-SLAM\cite{NGM-SLAM} and Constrained-SLAM\cite{Constrained} introduce multi-modal constraints (e.g., color, depth, scale) and hybrid loss backpropagation. Furthermore, systems like FGO-SLAM\cite{FGO-SLAM} and KBGS-SLAM\cite{KBGS-SLAM} trigger comprehensive history optimization upon loop closure, effectively eliminating residual drift by updating all historical poses and map points. By explicitly constraining scene geometry at pixel level, these methods enhance geometric accuracy throughout the map.

\begin{table}[]
\caption{ATE RMSE↓ [cm] on Various Datasets }
\label{tab4}
\centering
\scriptsize
\setlength{\tabcolsep}{3pt}
\begin{NiceTabular}{l|l|c|c|c|c}
    \toprule
      \textbf{Input} & \textbf{Methods}& \textbf{Optimization} & \textbf{Replica} & \textbf{TUM} & \textbf{ScanNet} \\
    \midrule
    \Block{8-1}{RGB}
      & GO-SLAM\cite{GO-SLAM}   & Global BA & 0.46 & 9.97 & 7.79 \\
      
      & NGM-SLAM\cite{NGM-SLAM}   & Global BA  & 8.51  & {--} & 8.05 \\
          & GI-SLAM\cite{GI-SLAM}   &  Local   & {--}  & 24.02 & {--} \\   
      & MGS-SLAM\cite{MGS-SLAM}   & Local    & 0.32  & 2.93 & {--} \\
      & Mon-SLAM\cite{Mon-SLAM}    &  Global Graph     & 0.32  & 5.37 & 7.34 \\
      & DROID-Splat\cite{DROID-Splat} &Global BA & 0.27 & \cellcolor{red!50}1.80 & {--} \\
        & SplatMAP\cite{SplatMAP}   & Local    & \cellcolor{red!50}0.18  & {--} & {--} \\

      & HI-SLAM2\cite{HI-SLAM2}    & Global BA   & 0.26  & {--} & \cellcolor{red!50}7.07 \\
    \midrule
    \Block{26-1}{RGB-D}
      & NeSLAM\cite{NeSLAM}  &   Local    & 0.66  & 2.01 & \cellcolor{yellow!30}6.98 \\
      & ESLAM\cite{ESLAM}   &  Local   & 0.62  & 2.34 & {--} \\
      & SplaTAM\cite{SplaTAM}     &  Local     & 0.36  & 3.81 & 12.76 \\
      & Loopy-SLAM\cite{Loopy-SLAM}&Global Graph& 0.29  & 3.85 & 7.70 \\
    & GLC-SLAM\cite{GLC-SLAM}    &  Global Graph  &0.23  & 2.23 & 9.20 \\
          & TlAMBRIDGE\cite{jiang2024tambridgebridgingframecenteredtracking}&Global BA & {--}  & 4.78 & {--} \\ 
     & DenseSplat\cite{DenseSplat}  &   Local   & 0.33  & 1.58 & 7.80 \\
      & Point-SLAM\cite{Point-SLAM} & Local & 0.52  & 8.92 & 12.72 \\

     & GI-SLAM\cite{GI-SLAM}   &  Local   & {--}  & 2.72 & {--} \\  
      & FT-SLAM\cite{FT-SLAM}    & Local   & {--}  & 1.40 & {--} \\  
      & NGM-SLAM\cite{NGM-SLAM}   & Global BA    & 0.51  & \cellcolor{yellow!30}1.04 & 7.27 \\ 
            & DROID-Splat\cite{DROID-Splat}& Global BA & 0.29  & 1.90 & {--} \\
    & Splat-SLAM\cite{Splat-SLAM}   & Global BA    &  0.34  & 2.10 & 7.50 \\ 
      & GS-Loop\cite{GS-Loop}&Global Graph & 0.28  & 1.49 & {--} \\ 
      & RTG-SLAM\cite{RTG-SLAM}    &  Local   & \cellcolor{yellow!30}0.18  & 1.06 & {--} \\

      & LoopSplat\cite{LoopSplat}   &  Global Graph     & 0.26  & 3.33 & 8.40 \\
         & OGS-SLAM\cite{OGS-SLAM} & Local & {--}  & 5.05 & {--} \\   
      & S3LAM\cite{S3LAM}& Local & 0.21   & 1.93   & {--} \\
            & MG-SLAM\cite{MG-SLAM}   &   Local    & 0.45  & {--} &\cellcolor{red!50} 6.77 \\
      & Constrained-SLAM\cite{Constrained}  &  Global BA    & 0.29  & 1.33 & 7.30 \\
    & KBGS-SLAM\cite{KBGS-SLAM}&Global BA & 0.27  & 2.55 & {--} \\
      & FGO-SLAM\cite{FGO-SLAM} &  Global BA & {--} &\cellcolor{orange!50} 0.98 & 7.37 \\
      & DSOSplat\cite{DSOSplat}  &  Local & 0.28  & {--} &\cellcolor{orange!50}6.80 \\
    & GSORB-SLAM\cite{GSORB-SLAM} & Local  & 0.38  & \cellcolor{red!50}0.91 & 9.32 \\
      & FGS-SLAM\cite{FGS-SLAM}    &  Local   & \cellcolor{orange!50}0.15  & 2.00 & {--} \\
      & GauS-SLAM\cite{GauS-SLAM}  &   Local    & \cellcolor{red!50}0.06  & 1.54 & 11.5 \\
    \bottomrule
    \end{NiceTabular}
    \end{table}

In summary, to improve the tracking accuracy of 3DGS-SLAM systems, researchers have conducted extensive studies across three complementary levels: local optimization, global graph optimization, and global BA.
Local optimization focuses on precise estimation within a single frame or a small local region, enhancing robustness through geometric and photometric constraints as well as adaptive Gaussian refinement.
Global graph optimization mitigates accumulated drift and error propagation by constructing keyframe graphs and incorporating loop closure detection, thereby improving overall consistency and scalability.
Building upon these, global BA optimization jointly refines camera poses and scene geometry by minimizing pixel-level residuals, explicitly constraining Gaussian representations to achieve high geometric consistency and accurate trajectory reconstruction. 

Table \ref{tab4} summarizes the tracking performance of representative methods on the Replica, TUM and ScanNet datasets. Fig.~\ref{6} compares tracking accuracy of representative methods on the Replica and TUM datasets.
Together, these approaches form a cohesive optimization pipeline that bridges local stability and global consistency, enabling 3DGS-SLAM systems to achieve higher accuracy in complex environments.

\subsection{Reconstruction Speed}
\label{3.3}
Reconstruction speed is a key metric for real-time SLAM performance. Faster mapping means quicker responsiveness and adaptability to dynamic environments. Recent research has optimized 3DGS-SLAM’s speed in three main areas: \textbf{Gaussian initialization}, \textbf{Gaussian densification}, and \textbf{parallel and hardware design}.

\textit{1)Gaussian Initialization Acceleration:}
3DGS-SLAM requires continuous iterative optimization of Gaussian properties to achieve better reconstruction quality. Better initialization results can reduce the number of algorithm iterations. To mitigate the computational cost of optimizing from random or sparse states, these methods leverage geometric priors and efficient sampling to achieve faster convergence with fewer iterations. Instead of starting from scratch, approaches like MGSO\cite{MGSO} and GPS-SLAM\cite{GPS-SLAM} utilize dense geometric priors—derived from DSO\cite{DSO} point clouds or Signed Distance Fields (SDF)—to directly initialize Gaussian positions and covariances, effectively bypassing the unstable early optimization phase. Similarly, to ensure rapid coverage without geometric priors, MemGS\cite{MemGS} employs a Patch-Grid sampling strategy that provides a more complete initial distribution, thereby accelerating the subsequent training convergence.

\textit{2)Gaussian Densification Acceleration:}
Gaussian densification entails the continuous creation, optimization, and update of a large number of primitives. The resulting explosion in Gaussian count dramatically increases the cost of both back-propagation and volume rendering, slowing both rendering and optimization. To address this, researchers propose selective optimization and hierarchical management to reduce redundancy. A common tactic is to restrict updates to essential primitives: RTG-SLAM\cite{RTG-SLAM} and MonoGS++\cite{MonoGS++} focus gradients on “unstable” regions or apply dynamic pruning to cull “floating” Gaussians, while FGS-SLAM\cite{FGS-SLAM} introduces a hierarchical scheme where only “core” Gaussians receive full-frequency updates. Beyond pruning, algorithmic efficiency is improved by reusing computed states or optimizing mathematical formulations. For instance, GS-ICP SLAM\cite{GS-ICP} recycles covariance matrices from tracking, GS-SLAM\cite{GS-SLAM} adopts a coarse-to-fine strategy to reduce resolution overhead, and CG-SLAM\cite{CG-SLAM} re-derives rasterization equations to optimize memory access patterns at the thread level. SAGA-SLAM\cite{SAGA-SLAM} further complements this by adaptively adjusting mapping strides based on feature density. These methods compress the Gaussian set or reduce redundant work, markedly increasing runtime speed.

\textit{3)Parallel and Hardware Design:}
To unlock the full parallel potential of 3DGS-SLAM and maximize system throughput, recent works optimize architecture through multi-threaded decoupling and hardware-specific acceleration to boost reconstruction speed. In terms of system architecture, frameworks like Photo-SLAM\cite{Photo} and SGR-SLAM\cite{SGR-SLAM} decouple tracking, mapping, and loop closure into asynchronous threads (often managing shared structures like super-voxel maps or octrees), preventing pipeline stalls. This is often paired with heterogeneous computing: RTG-SLAM\cite{RTG-SLAM} and SFGS-SLAM\cite{SFGS-SLAM} strategically offload frontend tracking to the CPU while reserving the GPU for intensive rendering and back-propagation. On the hardware level, custom accelerators and low-level optimizations are introduced to resolve memory bottlenecks. The KAIST team\cite{kaist} and GauSPU\cite{GauSPU} design specialized units for pixel reordering, symmetric reuse, and pipelined gradient updates to eliminate redundant computations. Similarly, GPS-SLAM\cite{GPS-SLAM} proposes a Gaussian-by-Gaussian parallelization strategy instead of a pixel-by-pixel approach (similar to CaRtGS\cite{CaRtGS}), avoiding conflicts in atomic operations. Additionally, it uses depth maps provided by the SDF for depth culling, thereby avoiding sorting.

\begin{figure}[!t] 
\centering 
\begin{tikzpicture}[>=stealth] 
\definecolor{color1}{rgb}{0.20, 0.70, 1.0} 
\definecolor{color2}{rgb}{0.10, 0.30, 0.60} 
\definecolor{color3}{rgb}{0.70, 0.87, 0.54} 
\definecolor{color4}{rgb}{0.89, 0.10, 0.11} 
\definecolor{color5}{rgb}{0.49, 0.70, 0.87} 
\definecolor{color6}{rgb}{0.8, 0.1, 0.7} 
\definecolor{color7}{rgb}{0.45, 0.25, 0.15} 
\definecolor{color8}{rgb}{0.99, 0.88, 0.60} 
\definecolor{color9}{rgb}{0, 0.39, 0} 
\definecolor{color10}{rgb}{0.80, 0.50, 0.00} 
\definecolor{color11}{rgb}{0.60, 0.20, 0.80} 
\definecolor{color12}{rgb}{0.00, 0.60, 0.60} 
\definecolor{color13}{rgb}{1, 0.75, 0.8} 
\definecolor{color14}{rgb}{0.40, 0.40, 0.40} 
\definecolor{color15}{rgb}{0.80, 0.20, 0.20} 
\definecolor{color16}{rgb}{0.20, 0.10, 0.40} 
\definecolor{color17}{rgb}{0.90, 0.50, 0.5} 
\definecolor{color18}{rgb}{0.70, 0.10, 0.2} 
\definecolor{color19}{rgb}{0.20, 0.90, 0.8} 
\definecolor{color20}{rgb}{0.0, 0.10, 0.7} 
\definecolor{color21}{rgb}{0.8, 0.60, 0.1} 
\definecolor{color22}{rgb}{0.4, 0.70, 0.3} 

\begin{axis}[
    name=mainaxis,
    xlabel={FPS $\uparrow$},
    ylabel={PSNR [dB] $\uparrow$},
    xlabel style={font=\footnotesize},
    ylabel style={font=\footnotesize},
    xmin=0, xmax=60,
    ymin=25, ymax=40,
    xtick={0,10,20,30,40,50,60},
    ytick={25,30,35,40},
    grid=major,
    grid style={dashed, gray!30},
    width=9cm,
    height=8cm,
    tick label style={font=\footnotesize},
    axis on top,
] 
\addplot[only marks, mark=*, color=color1] coordinates {(24.15, 34.33)}; 
\node[right, text=color1, font=\tiny] at (axis cs:24.15,34.33) {Orbeez-SLAM}; 

\addplot[only marks, mark=*, color=color18] coordinates {(0.30, 35.62)}; 
\node[right, text=color18, font=\tiny] at (axis cs:0.30,35.62) {Point-SLAM}; 

\addplot[only marks, mark=*, color=color3] coordinates {(0.23, 33.89)}; 
\node[right, text=color3, font=\tiny] at (axis cs:0.23,33.89) {SplaTAM}; 

\addplot[only marks, mark=*, color=color4] coordinates {(34.88, 34.96)}; 
\node[right, text=color4, font=\tiny] at (axis cs:34.88,34.96) {Photo-SLAM}; 

\addplot[only marks, mark=*, color=color5] coordinates {(8.34, 34.27)}; 
\node[right, text=color5, font=\tiny] at (axis cs:8.34,34.27) {GS-SLAM}; 

\addplot[only marks, mark=*, color=color6] coordinates {(15.4, 33.27)}; 
\node[right, text=color6, font=\tiny] at (axis cs:15.4,33.27) {CG-SLAM}; 

\addplot[only marks, mark=*, color=color7] coordinates {(30.00, 31.90)}; 
\node[right, text=color7, font=\tiny] at (axis cs:30.00,31.90) {MGSO}; 

\addplot[only marks, mark=*, color=color8] coordinates {(2.48, 37.79)}; 
\node[right, text=color8, font=\tiny] at (axis cs:2.48,37.79) {MonoGS++}; 

\addplot[only marks, mark=*, color=color9] coordinates {(17.24, 35.43)}; 
\node[right, text=color9, font=\tiny] at (axis cs:17.24,35.43) {RTG-SLAM}; 

\addplot[only marks, mark=*, color=color10] coordinates {(33.6, 34.00)}; 
\node[right, text=color10, font=\tiny] at (axis cs:33.6,34.00) {GauSPU}; 

\addplot[only marks, mark=*, color=color11] coordinates {(29.98, 38.83)}; 
\node[below, text=color11, font=\tiny] at (axis cs:29.98,38.83) {GSICP(limit)}; 

\addplot[only marks, mark=*, color=color12] coordinates {(10.57, 38.13)}; 
\node[right, text=color12, font=\tiny] at (axis cs:10.57, 38.13) {RD-SLAM}; 

\addplot[only marks, mark=*, color=color13] coordinates {(51.18, 29.21)}; 
\node[left, text=color13, font=\tiny] at (axis cs:51.18,29.21) {KAIST-SLAM}; 

\addplot[only marks, mark=*, color=color14] coordinates {(32.75, 38.75)}; 
\node[right, text=color14, font=\tiny] at (axis cs:32.75,38.75) {FGS-SLAM}; 

\addplot[only marks, mark=*, color=color15] coordinates {(34.67, 32.71)}; 
\node[right, text=color15, font=\tiny] at (axis cs:34.67, 32.71) {SGR-SLAM}; 

\addplot[only marks, mark=*, color=color16] coordinates {(29.97, 36.88)}; 
\node[right, text=color16, font=\tiny] at (axis cs:29.97, 36.88) {G2S-ICP}; 

\addplot[only marks, mark=*, color=color17] coordinates {(17.18, 38.75)}; 
\node[right, text=color17, font=\tiny] at (axis cs:17.18, 38.75) {SEGS-SLAM}; 

\addplot[only marks, mark=*, color=color2] coordinates {(30, 34.85)}; 
\node[above, text=color2, font=\tiny] at (axis cs:30, 34.85) {MemGS}; 

\addplot[only marks, mark=*, color=color19] coordinates {(33.17, 37.63)}; 
\node[right, text=color19, font=\tiny] at (axis cs:33.17, 37.63) {SFGS-SLAM}; 

\addplot[only marks, mark=*, color=color20] coordinates {(30, 39.13)}; 
\node[right, text=color20, font=\tiny] at (axis cs:30, 39.13) {GSICP+CaRtGS}; 

\addplot[only marks, mark=*, color=color21] coordinates {(60,35.93)}; 
\node[right, text=color21, font=\tiny] at (axis cs:60,35.93) {}; 

\addplot[only marks, mark=*, color=color22] coordinates {(60, 37.24)}; 
\node[right, text=color22, font=\tiny] at (axis cs:60, 37.24) {}; 

\pgfplotsextra{
    \draw[black, thick, solid]
        (axis cs:52, 35.3) rectangle (axis cs:60, 40); 
    \coordinate (mainAnchor) at (axis cs:56, 35.3); 
}
\end{axis} 

\begin{axis}[
    name=inset,
    at={(mainaxis.south west)},
    anchor=south west,
    xshift=1.0cm,
    yshift=0.6cm,
    width=5.6cm,
    height=3.6cm,
    xmin=90, xmax=270,  
    ymin=35, ymax=40,
    xtick={100,150,200,250},
    ytick={35,36,37,38,39,40},
    grid=major,
    grid style={dashed, gray!30},
    xlabel={},
    ylabel={},
    xlabel style={font=\tiny},
    ylabel style={font=\tiny},
    ticklabel style={font=\tiny},
    axis on top,
] 
\addplot[only marks, mark=*, mark size=2pt, color=color22] coordinates {(252.64, 37.24)}; 
\node[above left, font=\tiny, align=center, text=color22] at (axis cs:252.64,37.24) {\textbf{GPS-SLAM}\\252.6 FPS}; 

\addplot[only marks, mark=*, mark size=2pt, color=color21] coordinates {(98.11, 35.93)}; 
\node[below right, font=\tiny, align=center, text=color21] at (axis cs:98.11,35.93) {\textbf{GSICP(no limit)}\\(compact)}; 

\pgfplotsextra{
    \coordinate (insetAnchor) at (axis cs:270, 37); 
}
\end{axis}

\draw[->, thick, black, shorten >=2pt] 
  (mainAnchor) 
  |- ([yshift=-3cm, xshift=-0.5cm]mainAnchor) 
  -- (insetAnchor); 

\end{tikzpicture} 
\vspace{-0.5em}
\caption{Reconstruction speed on Replica dataset. An inset plot is included for methods that exceed the primary axis range.}
\label{7}
\end{figure}
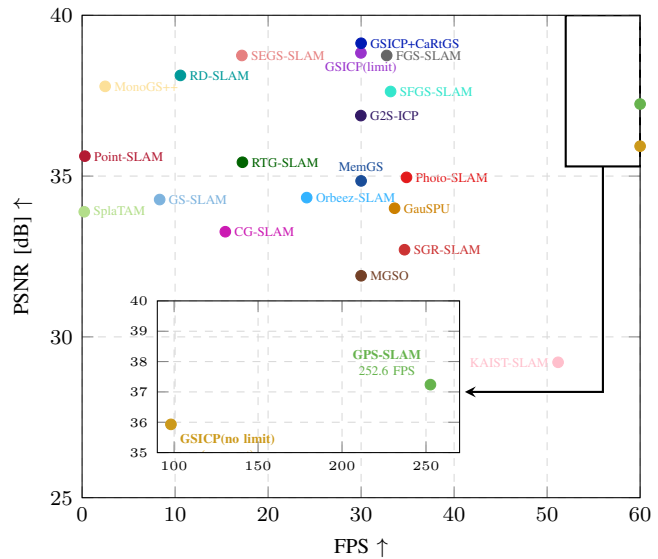

The performance improvement of 3DGS-SLAM systems has been driven by synergistic innovations across three key dimensions.
In the \textbf{initialization} stage, the integration of dense point cloud generators (e.g., DUSt3R\cite{DUSt3R}, MASt3R\cite{MASt3R}) and direct visual front-ends (e.g., DSO) significantly shortens early mapping time while providing high-quality initial 3D structure.
At the \textbf{representation} level, techniques such as Gaussian structure partitioning, adaptive insertion and pruning, and parameter compression reduce the number of Gaussian primitives and memory consumption, thereby alleviating the computational burden of rendering and optimization.
At the \textbf{computational strategy} level, multi-threaded architectures, customized CUDA kernels, and specialized hardware accelerators further improve system throughput and energy efficiency. Fig.~\ref{7} plots reconstruction speed (FPS) versus PSNR for various systems on the Replica dataset, and Table \ref{tab5} provides quantitative results.

The integration of these advances not only enhances the real-time performance and reconstruction accuracy of 3DGS-SLAM but also lays a solid foundation for its practical deployment on mobile and embedded platforms.

\begin{table}[]
\caption{Reconstruction Speed Evaluation on Replica Dataset }
\label{tab5}
\centering
\scriptsize
\setlength{\tabcolsep}{10pt}
\begin{NiceTabular}{l|c|c|c}
    \toprule
      \textbf{Methods} & \textbf{PSNR} $\boldsymbol{\uparrow}$ & \textbf{SSIM} $\boldsymbol{\uparrow}$ & \textbf{FPS} $\boldsymbol{\uparrow}$ \\
    \midrule
      
      Orbeez-SLAM \cite{Orbeez-SLAM}       & 14.33 & 0.768 & 24.15  \\
      Point-SLAM \cite{Point-SLAM}      & 35.62 & 0.970 & 0.30  \\
      SplaTAM \cite{SplaTAM}         & 33.89 & 0.970 & 0.23  \\
      Photo-SLAM \cite{Photo}      & 34.96 & 0.942 & \cellcolor{yellow!30}34.88  \\
      GS-SLAM \cite{GS-SLAM}         & 34.27   & 0.975   & 8.34  \\
      RD-SLAM \cite{RD-SLAM}          & 38.13 & 0.971 & 10.57 \\
      CG-SLAM \cite{CG-SLAM}  & 33.27   & {--}   & 15.40  \\
      MGSO \cite{MGSO}    & 31.90 & 0.910 & 30.00 \\
      MonoGS++ \cite{MonoGS++}        & 37.79 & 0.960 & 2.48  \\
      RTG-SLAM \cite{RTG-SLAM}          & 35.43 & \cellcolor{red!50}0.982 & 17.24 \\
      SGR-SLAM\cite{SGR-SLAM}& 32.71 & 0.930 & 34.67 \\
      GauSPU \cite{GauSPU}       & 34.00 & {--} & 33.6  \\
      MemGS \cite{MemGS} &34.85 & {--} & 30.00 \\
      GS-ICP(limit) \cite{GS-ICP}          & \cellcolor{yellow!30}38.83 & \cellcolor{yellow!30}0.975 & 29.98 \\
      GS-ICP(no limit) \cite{GS-ICP}          & 35.93 & 0.962 & \cellcolor{orange!50}98.11 \\
      GS-ICP+CaRtGS \cite{CaRtGS}          & \cellcolor{orange!50}39.19 & - & 30.00 \\
      G2S-ICP \cite{G2S-ICP}          &36.88 & 0.963 & 29.97 \\
      KAIST-SLAM \cite{kaist}       & 29.21 & 0.920 & 51.18  \\
      SFGS-SLAM\cite{SFGS-SLAM}& 37.63 & 0.972 & 33.17 \\
      SEGS-SLAM \cite{SEGS-SLAM}    & \cellcolor{red!50}39.42 & \cellcolor{orange!50}0.975 & 17.18  \\
      FGS-SLAM \cite{FGS-SLAM}       & 38.75 & 0.974 & 32.75  \\
      GPS-SLAM \cite{GPS-SLAM} & 37.24 & 0.960 & \cellcolor{red!50}252.64  \\

    \bottomrule
    \end{NiceTabular}
    \end{table}

\subsection{Memory Consumption}
\label{3.4}
\begin{figure*}[!t]
  \centering
  \includegraphics[width=\textwidth]{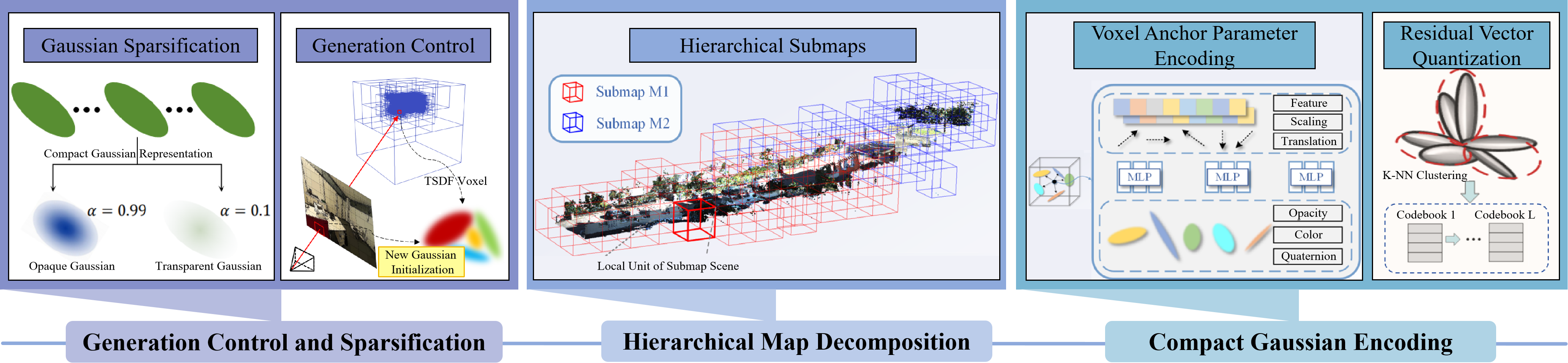} 
  \caption{Summary of memory consumption optimization methods. We categorize these strategies into three modules: Generation Control and Sparsification reduces redundancy by pruning insignificant Gaussians and limiting densification; Hierarchical Map Decomposition manages large-scale scenes by partitioning the map into scalable submaps; and Compact Gaussian Encoding compresses attributes via voxel-based anchoring and residual vector quantization.}
  \label{8}
\end{figure*}

Complex scenes can cause an explosion in the number of Gaussians, leading to high memory usage. Controlling map size without sacrificing quality and speed is a critical challenge. Existing research has pursued many strategies. As illustrated in Fig.~\ref{8}, we taxonomize memory optimization strategies into three complementary classes: \textbf{Gaussian generation control and sparsification}, \textbf{hierarchical map decomposition}, and \textbf{compact Gaussian encoding}, and we detail their guiding concepts, pivotal techniques, and mutual interplay.

\textit{1) Gaussian Generation Control and Sparsification:} In large-scale reconstructions, the unconstrained proliferation of Gaussian primitives often leads to prohibitive memory consumption. To address this, current approaches implement strict spatial constraints and adaptive pruning strategies, aiming to strike a balance between reconstruction fidelity and storage efficiency. Proactive generation control serves as the first line of defense by preventing redundancy at the source. Instead of initializing Gaussians indiscriminately, methods like RTG-SLAM\cite{RTG-SLAM}, GS-Fusion\cite{GSFusion}, and 2DGS-SLAM\cite{2DGS-SLAM} enforce rigorous occupancy checks. By utilizing surface opacity labels, TSDF-guided quadtree structures, or voxel hash tables to verify spatial occupancy, these systems ensure that new primitives are spawned only in strictly unmapped or visible regions, thereby avoiding the overlap of redundant Gaussians. In parallel, information-theoretic sampling strategies optimize the initial distribution of primitives. CompactGS\cite{CompactGS} and MGSO\cite{MGSO} move away from uniform initialization, instead leveraging geometric priors (such as DSO point clouds) or image gradients to guide placement. This allows for dense clustering in geometrically complex areas to capture fine details, while maintaining a sparse representation in textureless or flat regions. Complementing these generation policies, reactive pruning and merging mechanisms dynamically refine the map by eliminating unnecessary primitives post-creation. MotionGS\cite{MotionGS} and GPS-SLAM\cite{GPS-SLAM} incorporate sparsity losses or SDF-based constraints into the optimization objective, automatically penalizing and filtering out low-opacity or insignificant Gaussians. Furthermore, MemGS\cite{MemGS} addresses geometric redundancy by calculating the Mahalanobis distance between neighboring primitives, merging those with high similarity into a single representation to maintain map compactness without sacrificing quality.

\textit{2) Hierarchical Map Decomposition:} As the scale of the reconstructed environment grows, maintaining a monolithic global map becomes computationally intractable due to unbounded memory growth. To avoid this, these approaches distribute the scene into manageable submaps or subgraphs, enabling on-demand activation and optimization. Frameworks like VPGS-SLAM\cite{VPGS-SLAM} and DenseSplat\cite{DenseSplat} logically partition the scene based on camera motion thresholds or fixed frame intervals. This allows the system to keep only the locally relevant submaps active while inactive regions are “put to sleep”, effectively decoupling map size from real-time performance. NGM-SLAM\cite{NGM-SLAM} takes this a step further by employing hybrid “neural submaps”: it represents local scenes using lightweight NeRF modules and only falls back to explicit Gaussian rendering when strictly necessary, thereby reducing the overhead of maintaining millions of explicit primitives. Furthermore, efficient fusion mechanisms ensure linear memory scaling during loop closure or merging:  VPGS-SLAM\cite{VPGS-SLAM} applies an online distillation process during submap fusion, effectively compressing the knowledge from overlapping submaps into a unified representation. Similarly, DenseSplat\cite{DenseSplat} and NGM-SLAM\cite{NGM-SLAM} implement aggressive pruning strategies and two-stage optimization pipelines. By identifying and removing redundant Gaussians before integrating local submaps into the global frame, these methods ensure that memory usage scales linearly rather than exponentially with scene exploration.

\textit{3) Compact Gaussian Encoding:} The standard 3D Gaussian representation requires storing high-dimensional attributes for millions of primitives, leading to a massive memory footprint. To counter this, recent methods pursue dimensionality reduction and implicit encoding to fundamentally lower the storage cost per primitive. Attribute compression is a direct approach, which reduces the bit-width of stored parameters. CGS-SLAM\cite{CGS-SLAM} utilizes Residual Vector Quantization (R-VQ) to map continuous parameters (rotation, scale, color) to discrete codebooks, while MGSO\cite{MGSO} simplifies the representation by substituting spherical harmonics with raw RGB values. Alternatively, structural re-parameterization reduces storage requirements by altering the fundamental representation. For instance, VPGS-SLAM \cite{VPGS-SLAM} introduces a memory-efficient voxel-anchoring scheme: rather than storing explicit parameters for every Gaussian, it partitions space into a sparse voxel grid, where each “anchor” voxel predicts parameters via a lightweight MLP. This allows the system to decode Gaussian properties on-the-fly, effectively replacing static storage with implicit neural computation. Similarly, S3LAM \cite{S3LAM} replaces volumetric Gaussians with 2D surfels, leveraging a surfel-based management strategy to achieve a more memory-efficient geometric representation.

\begin{figure*}[!t]
    \centering
\includegraphics[scale=0.65]{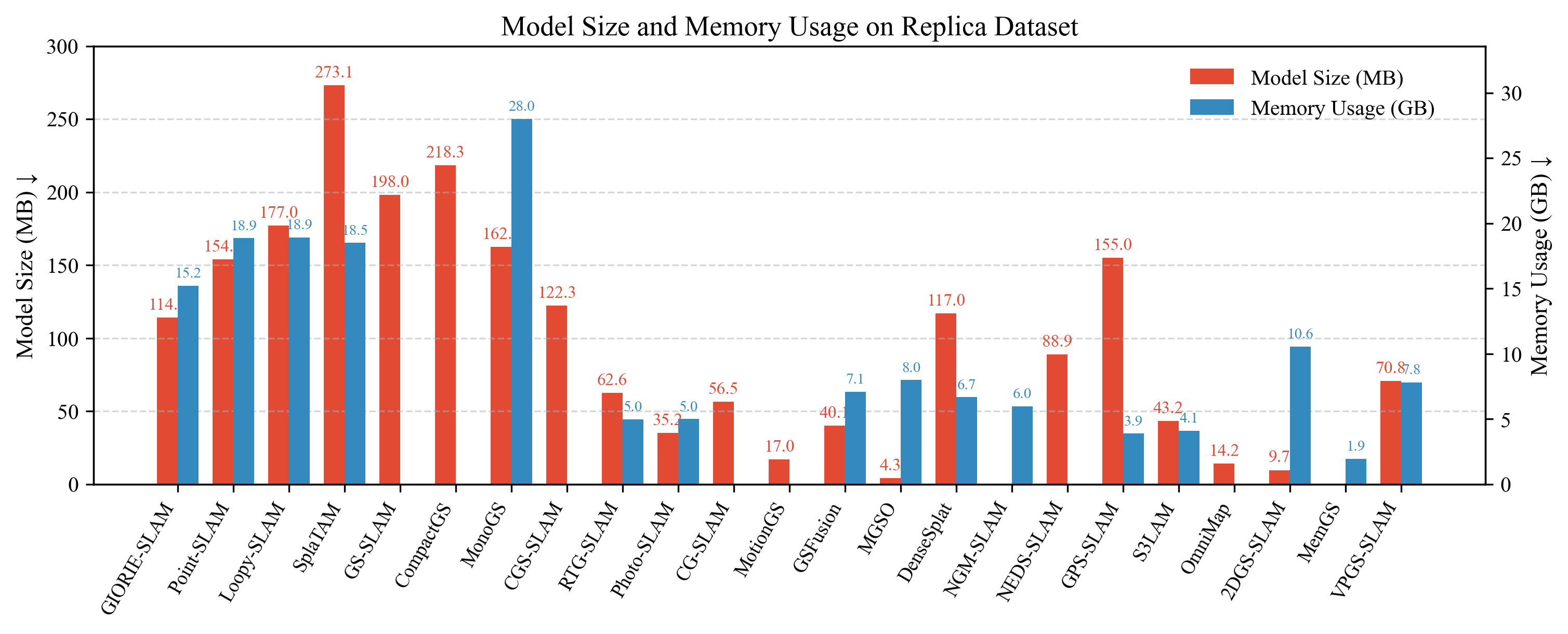}
    \caption{Memory consumption comparison on the Replica dataset. “Model size” indicates the stored map. }
    \label{9}
\end{figure*}

\begin{table}[]
  \centering
  \caption{Memory Consumption Evaluation on Replica Dataset}
  \label{tab6}
\scriptsize
\begin{NiceTabular}{l|c|c}
  \toprule
  \textbf{Methods} & \textbf{Model size(MB)}$\boldsymbol{\downarrow}$ & \textbf{Memory usage(GB)}$\boldsymbol{\downarrow}$ \\
  \midrule
  GIORIE-SLAM \cite{GIORIE-SLAM} & 114.00 & 15.22 \\ 
  Point-SLAM  \cite{Point-SLAM}& 154.00 & 18.86 \\ 
  Loopy-SLAM  \cite{Loopy-SLAM}& 177.00 & 18.91 \\ 
  SplaTAM \cite{SplaTAM}& 273.09 & 18.50 \\ 
  GS-SLAM\cite{GS-SLAM}& 198.04 & - \\ 
  CompactGS\cite{CompactGS}& 218.35 & - \\ 
  MonoGS  \cite{MonoGS}& 162.41 & 27.99 \\ 
  CGS-SLAM\cite{CGS-SLAM}& 122.29& - \\
  RTG-SLAM  \cite{RTG-SLAM}& 62.60 & 4.99 \\ 
  Photo-SLAM  \cite{Photo}& 35.21 & 5.00 \\ 
  CG-SLAM\cite{CG-SLAM}& 56.50& - \\
  MotionGS  \cite{MotionGS}& 17.00 & - \\ 
  GSFusion  \cite{GSFusion}& 40.10 & 7.08 \\ 
  MGSO  \cite{MGSO}& \cellcolor{red!50}4.30 & 7.98 \\ 
  DenseSplat \cite{DenseSplat}& 117.00 & 6.67 \\ 
  NGM-SLAM \cite{NGM-SLAM}& - & 5.98 \\ 
  NEDS-SLAM\cite{NEDS-SLAM}& 88.93 & - \\ 
  GPS-SLAM \cite{GPS-SLAM}& 155.00 & \cellcolor{orange!50}3.88 \\ 
  S3LAM\cite{S3LAM}& 43.20 & \cellcolor{yellow!30}4.10 \\ 
  OmniMap\cite{OmniMap}& \cellcolor{yellow!30}14.20 & - \\
  2DGS-SLAM\cite{2DGS-SLAM}& \cellcolor{orange!50}9.70 & 10.56 \\ 
  MemGS\cite{MemGS}& - & \cellcolor{red!50}1.95 \\ 
  VPGS-SLAM \cite{VPGS-SLAM}& 70.81 & 7.80 \\ 
  \bottomrule
  \end{NiceTabular}
  \end{table}
In summary, to address the memory overhead caused by the rapid expansion of Gaussian representations, recent 3DGS-SLAM systems have established a multilayer optimization pipeline spanning generation control, structural management, and representation compression.
\textbf{Gaussian generation control and sparsification} effectively suppress redundant Gaussian initialization, while \textbf{hierarchical map decomposition} distribute memory usage through localized scheduling. \textbf{Lightweight encoding techniques} further reduce per-Gaussian storage costs by compressing parameter representations.
Several works integrate these strategies; for instance, VPGS-SLAM~\cite{VPGS-SLAM} merges redundant Gaussians after loop closure, partitions the global map into multiple submaps to prevent memory duplication, and employs neural encoding for compact Gaussian representation.

Although each category focuses on distinct optimization objectives, together they enable efficient, compact, and high-fidelity map representations for large-scale 3DGS-SLAM systems.
Table \ref{tab6} reports memory usage on the Replica dataset and Fig.~\ref{9} visualizes the comparison. In summary, by combining intelligent generation control, map decomposition, and compact encoding, 3DGS-SLAM systems effectively mitigate the memory explosion challenge.

\section{Robustness Enhancements in Complex Environments}
Recent advances in 3DGS-SLAM have achieved remarkable progress in performance optimization. Beyond performance, robustness to challenging conditions is crucial. However, most existing optimization efforts are conducted under static or quasi-static assumptions, leaving system robustness severely tested in dynamic and unpredictable environments. When exposed to rapid camera motion or the presence of moving objects, 3DGS-SLAM systems face severe challenges—motion blur disrupts feature extraction and tracking, while dynamic elements introduce inconsistency into static scene reconstruction. These factors significantly degrade performance and stability in complex real-world scenarios. To address these issues, recent studies have explored robustness-oriented strategies that specifically target \textbf{motion blur}(Sec \ref{4.1}) suppression and \textbf{dynamic scenes}(Sec \ref{4.2}) adaptation.
This section reviews representative works along these two directions and analyzes their underlying principles and effectiveness.

\subsection{Motion Blur}
\label{4.1}
Motion blur occurs when the camera moves rapidly during exposure, stretching textures and smearing edges\cite{ASSID}. In SLAM, blur degrades feature extraction and matching, reducing geometric accuracy and destabilizing pose estimation and mapping. Under low light or fast motion\cite{CAMO-MOT}, blur often causes tracking drift, map inconsistency, or even failure. Traditional SLAM methods handle blur with pre-filtering (DeblurSLAM\cite{DeblurSLAM}), blur-aware VO (MBA-VO\cite{9710590}), event cameras (EN-SLAM\cite{10655500}), or semantic segmentation\cite{s25061696}. 

\begin{figure}[!t]
    \centering
    \includegraphics[width=0.48\textwidth]{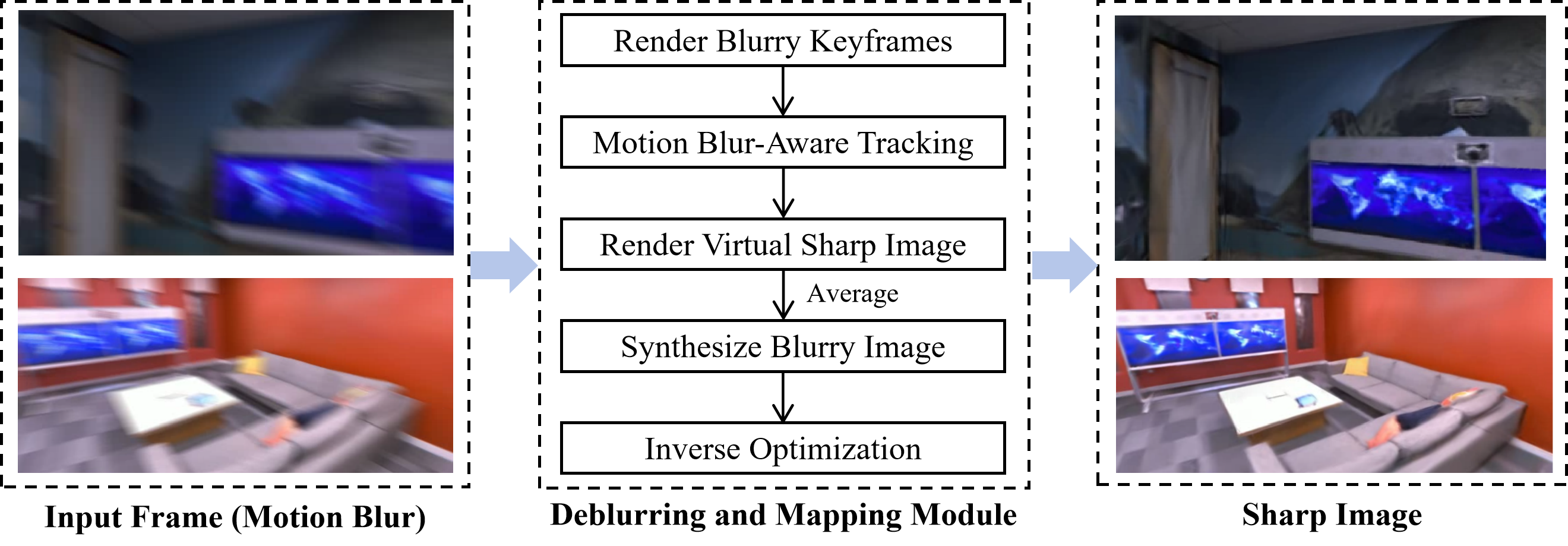}
    \caption{Motion blur optimization framework.}
    \label{10}
\end{figure}

In the context of 3DGS-SLAM, this challenge is exacerbated because Gaussian initialization and splitting rely heavily on sharp photometric cues; blurred inputs often lead to “floater” artifacts, tracking drift, or map inconsistency. Recent advances in 3DGS-SLAM have yielded several representative approaches that significantly improve system robustness and mapping accuracy in the presence of motion blur. These methods can be broadly categorized into two paradigms: robust system coupling, which tightly integrates front-end perception and back-end optimization to jointly mitigate motion-induced degradation, and explicit physical modeling, which directly accounts for motion-induced image formation effects—such as trajectory-aware rendering—to recover more accurate scene representations. Fig.~\ref{10} presents the overall framework of the motion-blur optimization method.

\textit{1) Robust System Coupling:} Instead of explicitly modeling the blur kernel, some approaches focus on enhancing system robustness against poor-quality data through tighter frontend-backend coupling. TAMBRIDGE\cite{jiang2024tambridgebridgingframecenteredtracking} addresses the disconnect between tracking and mapping in blurred scenarios by introducing a “fusion bridge”. By selecting optimal key views and jointly optimizing reprojection errors, it suppresses convergence instability caused by blur or occlusion. Furthermore, it incorporates a boundary mask and residual fusion mechanism to filter out unreliable regions, ensuring that the Gaussian map remains coherent even when the input sequence contains extended periods of blur.

\textit{2) Explicit Physical Modeling:} Diverging from filtering or masking strategies, other works aim to fundamentally solve the problem by mathematically modeling the image formation process of motion blur. MBA-SLAM\cite{wang2024mbaslammotionbluraware} proposes an end-to-end blur-aware framework that models the camera’s motion during exposure as a continuous SE(3) trajectory. Based on this trajectory, it synthesizes a “reblurred” image from the 3D Gaussians to align with the blurred observation. By jointly optimizing the scene geometry and the motion trajectory, MBA-SLAM effectively turns the blur from an artifact into a geometric constraint, allowing the system to utilize blurred frames for accurate tracking.

Sharing a similar motivation but employing a discrete temporal formulation, Deblur-SLAM\cite{9483818} models the blurred frame as an integration of multiple “virtual clear subframes”. The system generates these virtual images via interpolation and averages them to approximate the real blurred input. It then minimizes the photometric and geometric errors between the synthesized average and the actual observation. When combined with online loop detection and global Bundle Adjustment (BA), this approach allows for the recovery of sharp, high-quality maps from severely blurred data by effectively decomposing the blur integral.

These methods use blur modeling, trajectory interpolation, reblurred rendering, and multi-scale alignment to achieve robust tracking and mapping in blurred conditions. They greatly improve 3DGS-SLAM’s pose stability and map clarity in motion-blurred scenes. Future work may further integrate blur modeling with depth cues, semantic understanding, and multimodal sensors (e.g., events or inertial) to enhance generalization and robustness to extreme blur.

\subsection{Dynamic Scenes}
\label{4.2}
Dynamic scenes contain moving objects, violating the static world assumption. In such environments, conventional SLAM methods can mistake dynamic features for the static background, leading to errors in feature matching, pose estimation, and map construction. These errors can accumulate over time, causing trajectory drift, loop-closure failure, or even complete map breakdown. To address this, a range of robust SLAM frameworks apply techniques like semantic segmentation masking (DynaSLAM\cite{DynaSLAM}, NID-SLAM\cite{NID-SLAM}), foreground–background separation (DDN-SLAM\cite{DDN-SLAM}), and residual-based feature filtering (RoDyn-SLAM\cite{RoDyn-SLAM}, SD-SLAM\cite{Semantic-independent}) . 

In 3DGS-SLAM, moving objects can create redundant Gaussians\cite{Motion-awareGS} and wrong color/depth cues\cite{FRPGS}, further degrading results. To mitigate these effects, emerging approaches have proposed some methods to improve motion awareness and dynamic suppression. The following introduces several representative methods that address these challenges from distinct technical perspectives, providing critical support for stable deployment of 3DGS-SLAM in complex real-world environments. Table \ref{tab7} categorizes these methods. Each group’s advantages and limitations are summarized, along with typical application scenarios. 

\begin{table*}
\centering
\caption{Summary of Dynamic Scene Handling Methods in 3DGS-SLAM}
\label{tab7}
\footnotesize
\begin{tabular*}{\textwidth}{
  @{\extracolsep{\fill}}
  p{1.3cm}      
  p{4.4cm} 
  p{3cm} 
  p{4.5cm} 
  p{3.1cm}
  @{\extracolsep{\fill}}
}
\toprule
\textbf{Category} & \textbf{Representative Methods} & \textbf{Advantages} & \textbf{Limitations} & \textbf{Typical Scenarios}\\
\midrule

\multirow{2}{=}{Semantic\\Priors} 
& \multirow{2}{=}{DG\cite{DG-SLAM}, DGS\cite{DGS-SLAM}, DyPho\cite{DyPho-SLAM},  SDD\cite{SDD-SLAM}, DyGS\cite{DyGS-SLAM},Go\cite{Goo-SLAM},}
& \multirow{2}{=}{Object-level mapping; Semantically aware.} 
& \multirow{2}{=}{High compute cost; Fails on unseen classes; Generalization-dependent.} 
& \multirow{2}{=}{Structured scenes with known object categories.} \\ \\ 
\midrule
\multirow{2}{=}{Geometric\\Consistency}
& \multirow{2}{=}{Dy3DGS-SLAM\cite{Dy3DGS-SLAM}, GARAD-SLAM\cite{li2025garadslam3dgaussiansplatting}, Gassidy\cite{Gassidy}}
& \multirow{2}{=}{Prior-free; Handles unknown objects.} 
& \multirow{2}{=}{Coarse dynamic geometry; Sensitive to fast motion \& view changes.} 
& \multirow{2}{=}{Large-scale scenes with unknown dynamics.} \\ \\ 
\midrule
\multirow{3}{=}{Explicit\\Dynamic\\Modeling}
& \multirow{3}{=}{ADD-SLAM\cite{ADD-SLAM}, PG-SLAM\cite{PG-SLAM}, DynaGSLAM\cite{DynaGSLAM}}
& \multirow{3}{=}{Full dynamic reconstruction; Decoupled FG/BG optimization.} 
& \multirow{3}{=}{High memory/compute usage; Relies on segmentation \& tracking.} 
& \multirow{3}{=}{Interactive or slow-moving dynamic tasks.} \\ \\ \\ 
\midrule
\multirow{2}{=}{Uncertainty\\Modeling}
& \multirow{2}{=}{UP-SLAM\cite{UP-SLAM}, WildGS-SLAM\cite{zheng2025wildgsslammonoculargaussiansplatting}}
& \multirow{2}{=}{Threshold-free; Flexible structure; Robust.} 
& \multirow{2}{=}{Low interpretability; Complex training; Feature-dependent.} 
& \multirow{2}{=}{Scenes with missing labels or unknowns.} \\ \\ 

\bottomrule
\end{tabular*}
\end{table*}

\textit{1) Semantic Prior Methods:}
Leveraging the semantic reasoning capabilities of modern foundation models, these approaches leverage advanced vision–language models (VLMs) such as Mask R-CNN, and SAM, together with large language models (LLMS) for label generation, to track and exclude dynamic objects belonging to known semantic categories.  Advanced segmentation integration forms the baseline: Go-SLAM\cite{Goo-SLAM} combines ChatGPT-4o and VLMs to generate open-vocabulary masks, assigning unique semantic IDs to prevent dynamic fusion. However, raw semantic predictions often suffer from boundary inaccuracies or temporal flicker. To address this, recent works introduce spatio-temporal refinement mechanisms. DyPho-SLAM\cite{DyPho-SLAM}, DGS-SLAM\cite{DGS-SLAM}, and DG-SLAM\cite{DG-SLAM} leverage multi-frame consistency—utilizing static background priors, residual histograms, or reprojection error analysis—to robustly identify and suppress dynamic outliers that deviate from the stable map. Focusing on boundary precision, SDD-SLAM\cite{SDD-SLAM} and DyGS-SLAM\cite{DyGS-SLAM} correct coarse semantic masks by aligning them with depth discontinuities or refining bounding boxes via clustering and Gaussian Mixture Model (GGM), ensuring that geometric edges match semantic labels. Furthermore, SDD-SLAM extends this logic to passive dynamics, tracking the semantic center shifts of objects to identify and remove typically static items that strictly motion-based methods might miss.

\textit{2) Geometric Consistency Methods:} In absence of semantic labels, these methods identify dynamic pixels by detecting discrepancies between input frames and static reconstruction. A common approach is motion-based penalization: Dy3DGS-SLAM\cite{Dy3DGS-SLAM} utilizes optical flow and monocular depth to generate probabilistic motion masks, applying additional losses to penalize Gaussians associated with moving regions. Instead, statistical and probabilistic modeling is used to distinguish dynamics: Gassidy\cite{Gassidy} segments each input frame into potential object and background regions via instance segmentation and employs a GGM to classify regions based on photometric and geometric loss behaviors, while GARAD-SLAM\cite{li2025garadslam3dgaussiansplatting} builds a Conditional Random Field over Gaussian attributes to label dynamics, validating the segmentation via sparse flow. These methods offer adaptability to unknown objects but may struggle with fine details or rapid viewpoint changes.

\textit{3) Explicit Dynamic Modeling:} Rather than treating moving objects simply as outliers to be removed, these approaches aim to decouple and reconstruct dynamic elements alongside the static background. Model-based reconstruction is employed when specific prior knowledge is available: PG-SLAM\cite{PG-SLAM} utilizes human-shape priors (SMPL) to constrain deformation, jointly rendering background and foreground to estimate robust poses. Conversely, for general moving objects where pre-built models are unavailable, joint estimation and multi-stream representation are adopted. ADD-SLAM\cite{ADD-SLAM} maintains separate Gaussian sequences for each object, estimating their motion online through dynamic-rendering losses. DynaGSLAM\cite{DynaGSLAM} further advances this philosophy by explicitly critiquing removal-based methods; instead of discarding moving regions, it incorporates a motion prediction module to jointly estimate accurate ego-motion and the trajectories of dynamic objects. This allows the system to achieve high-quality, real-time rendering of the full dynamic scene—both static background and moving entities—rather than leaving “holes” where dynamic objects once stood.

\textit{4) Uncertainty Modeling:} Departing from binary masks, these methods adopt a probabilistic perspective, using neural networks to infer pixel-wise uncertainty and down-weight unreliable regions during optimization. Feature-based uncertainty estimation is central to this strategy: UP-SLAM\cite{UP-SLAM} and WildGS-SLAM\cite{zheng2025wildgsslammonoculargaussiansplatting} decode high-level features (from DINO or DINOv2) via MLPs to predict uncertainty maps. These maps serve as adaptive weighting factors: UP-SLAM fuses temporal and visual features to enhance robustness, while WildGS-SLAM integrates uncertainty into the DBA\cite{DROID_SLAM} tracking backend and mapping loss, effectively suppressing the influence of dynamic Gaussians without explicit segmentation. While flexible, these neural approaches require careful training to ensure interpretability.
\begin{table}[]
\caption{Average ATE and STD Results on Bonn and TUM Dataset}
\label{tab8}
\centering
\scriptsize
\setlength{\tabcolsep}{2pt}
\begin{NiceTabular}{l|cc|cc}
\noalign{\hrule height 1pt}
                          & \multicolumn{2}{c|}{Bonn} & \multicolumn{2}{c}{TUM} \\ \cline{2-5} 
\multirow{-2}{*}{Methods} & \textbf{ATE (Avg.)}$\downarrow$ & \textbf{STD (Avg.)}$\downarrow$   & \textbf{ATE (Avg.)}$\downarrow$ & \textbf{STD (Avg.)}$\downarrow$ \\ \hline
    ORB-SLAM3\cite{ORB-SLAM3}   & 62.84  & 32.86& 11.10 & 4.30 \\ 
    DROID-SLAM\cite{DROID_SLAM}  & 15.40 & {--}& {--} & {--} \\ 
    NID-SLAM\cite{NID-SLAM}    & 10.80 & 6.93 & 18.61 & 12.08 \\
    DynaSLAM\cite{DynaSLAM}    & 4.74 & {--} & 2.00 & {--} \\   
    DDN-SLAM\cite{DDN-SLAM}   & 3.00 & 1.60 & 2.05 & 1.24 \\ 
    SplaTAM\cite{SplaTAM}     & 125.94 & 62.25& 166.00 & 52.90\\
    Photo-SLAM\cite{Photo}  & 62.79 & 31.71 & 34.18 & 17.59\\
    GS-ICP SLAM\cite{GS-ICP} & 49.20 & 20.50 & 42.90 & 18.40\\
    RoDyn-SLAM\cite{RoDyn-SLAM} & 12.10 & 4.32 & 4.10 & 2.30\\   
    DG-SLAM\cite{DG-SLAM}  & 5.51 & 2.79 & 2.20 & {--}\\
    DGS-SLAM\cite{DGS-SLAM}    & 10.75 & {--} & 4.61 &{--}\\
    Gassidy\cite{Gassidy}     &7.80 & 3.10 &  2.60 & 1.30 \\
    Dy3DGS-SLAM\cite{Dy3DGS-SLAM} & 4.50 & {--}& 4.70 &{--} \\
    DyPho-SLAM\cite{DyPho-SLAM}    &  {--} &  {--} & 1.60 &  \cellcolor{yellow!30}0.70\\
    SDD-SLAM\cite{SDD-SLAM}    &  3.77 &  {--} & 1.80 &{--}\\
    PG-SLAM\cite{PG-SLAM}    & 6.50 & 2.20  & 4.50 & 1.80 \\
    DyGS-SLAM\cite{DyGS-SLAM} & 3.10 & {--} & 1.80 & {--} \\
    UP-SLAM\cite{UP-SLAM}     & 3.20 & {--} &  \cellcolor{yellow!30}1.42 &{--}\\
    WildGS-SLAM\cite{zheng2025wildgsslammonoculargaussiansplatting} & \cellcolor{yellow!30}2.88 & \cellcolor{yellow!30}1.45 &\cellcolor{orange!50} 1.32 &\cellcolor{orange!50} 0.67 \\

    ADD-SLAM\cite{ADD-SLAM}   & \cellcolor{orange!50}2.77 & \cellcolor{red!50}1.05 & \cellcolor{red!50}1.25 &\cellcolor{red!50} 0.65 \\ 
    GARAD-SLAM\cite{li2025garadslam3dgaussiansplatting} & \cellcolor{red!50}2.68 & \cellcolor{orange!50}1.22& 1.94& 1.15 \\ 
\hline
  \end{NiceTabular}
\end{table}

Overall, 3DGS-SLAM systems have evolved from passive adaptation under static-world assumptions to robust mapping paradigms that actively integrate dynamic recognition, structural disentanglement, and adaptive optimization. Future research may focus on cross-frame motion consistency modeling, self-supervised dynamic pattern discovery, and multimodal perception fusion. Table \ref{tab8} summarizes the tracking performance of representative algorithms on the Bonn dataset.

\section{Future Research Directions}
We highlight promising directions for advancing 3DGS-SLAM in this section.
\subsection{Event-Camera-Based Blur Handling}
Current blur-aware methods in 3DGS-SLAM separate deblurring and Gaussian optimization, usually assuming linear blur models. This decoupling limits performance: extensive image generation and residual computation reduce real-time speed, and non-linear or non-rigid blurs still pose problems. 

A future direction is to integrate event cameras into 3DGS-SLAM. Event cameras provide microsecond-resolution, high-dynamic-range asynchronous data, capturing continuous motion information even under extreme motion or lighting. Recently, several studies have attempted to introduce event cameras into 3DGS\cite{EvaGaussians,E2GS,E-3DGS,DiET-GS,EventSplat,SweepEvGS}. From this, by fusing events with RGB images in an end-to-end system, one could achieve robust SLAM in high speed, or strongly blurred scenarios. Designing a unified blur model and a framework that simultaneously fuses full-frame images and events would allow 3DGS-SLAM to operate reliably in conditions that defeat traditional cameras.

\subsection{Reconstruction in Extreme Environments}
3DGS-SLAM excels in standard indoor or outdoor scenes, but extreme conditions remain challenging. In texture-sparse or highly repetitive scenes (e.g., snow, desert, fog), images provide limited information. Gaussian initialization lacks constraints, harming map quality and speed. In outdoor terrains, limited viewpoints (due to obstacles or terrain) lead to unobservable geometry and sparse points. Dust or rain can confuse keyframe selection by masking the background as obstacles. 

Future work should address these with multi-modal perception, prior driven mapping, and anti-interference strategies. For example, adding other sensors can compensate for poor RGB data in low-texture or low-light conditions. Learning-based priors or generative models could infer unseen scene geometry from limited views, improving map continuity. Robust frame selection and masking schemes (using temporal consistency or learned occlusion detectors) could avoid false tracking cues in dusty/rainy conditions. Combining these strategies may allow 3DGS-SLAM to generalize to harsh real-world scenarios, gradually closing the perception gap of conventional systems.

\subsection{Incorporating Physical Attributes}
To date, 3DGS-SLAM focuses on geometry and appearance, assuming static or quasi-rigid scenes. Real-world objects, however, have physical behaviors and time-varying properties. Early 3DGS research\cite{PUGS,RainyGS} has started to encode simple physics and deformable objects. 

Future research could systematically integrate physical attributes into 3DGS-SLAM. For example, physics simulators or learned physical field supervision could teach the 3DGS model about elasticity, density, or friction. Gaussians could be augmented with physical state (velocity, material parameters), enabling fine reconstruction and prediction of non-rigid objects (e.g., cloth, fluids). Mechanics-based constraints could improve modeling of complex interactions. Physically-based rendering extensions could recover material properties from appearance. Enhancing 3DGS-SLAM with physics awareness would be valuable for robotics, AR and other tasks requiring physical scene understanding.

\subsection{Integration with Large Vision Models}
Current 3DGS-SLAM relies on classical SLAM frameworks and geometric priors, which struggle in low-texture or structureless environments. Concurrently, emerging large-scale vision models (e.g., Transformers) enable self-supervised, end-to-end learning of cross-view geometry and camera motion with minimal supervision. For example, Visual Geometry Grounded Transformer (VGGT\cite{VGGT}) models can achieve high-quality reconstruction without intrinsic calibration or IMU\cite{ReviewVGGT}, demonstrating robustness beyond traditional pipelines. 

Some works\cite{VGGT-SLAM,VGGT-Long} have begun to embed such models into SLAM. Future 3DGS-SLAM could further integrate these large models: using their end-to-end learned features to improve frontend robustness and adaptability, while relying on 3DGS for efficient explicit mapping. These models could also be extended for tasks like multi-view fusion, temporal context encoding, and dynamic object disentanglement. They could guide Gaussian initialization, generate dynamic masks, and provide scene understanding. Marrying 3DGS-SLAM with foundation models promises to imbue the system with greater generality and learning capability across diverse environments.

\section{Conclusion}
This survey has provided a comprehensive review of research at the intersection of 3DGS and SLAM. We have detailed how 3DGS-SLAM systems achieve high-fidelity and efficient mapping, examining the key optimizations and robustness strategies that drive next-generation SLAM performance. By systematically organizing advances in rendering quality, tracking accuracy, reconstruction speed, memory consumption, and robustness, we have highlighted the multi-dimensional progress in this field. Looking forward, emerging technologies—such as event-based sensing, physics-aware modeling, and large-scale vision models—offer exciting avenues to further enhance 3DGS-SLAM. We hope this survey serves as a foundation for researchers to build more capable and robust SLAM systems for complex real-world applications.

\bibliographystyle{IEEEtran}        
\bibliography{bibtex}           

\vfill

\end{document}